\documentclass[lettersize,journal]{IEEEtran}
\usepackage{amsmath,amsfonts}
\usepackage{algorithmic}
\usepackage{algorithm}
\usepackage{array}
\usepackage[caption=false,font=footnotesize,labelfont=rm,textfont=rm]{subfig}
\usepackage{textcomp}
\usepackage{stfloats}
\usepackage{url}
\usepackage{verbatim}
\usepackage{graphicx}
\usepackage{svg}
\usepackage{pdfpages}
\usepackage{cite}
\usepackage{booktabs}
\usepackage{multirow}
\usepackage{ulem}
\usepackage[justification=centering,font=footnotesize]{caption}
\usepackage{paralist}
\usepackage{ulem}
\usepackage{microtype}
\usepackage{arydshln}
\usepackage{booktabs}
\usepackage{dashrule}

\begin{document}

\title{Filter Pruning based on Information Capacity and Independence}
\author{Xiaolong Tang, 
			Shuo Ye,
			Yufeng Shi,
			Tianheng Hu,
			Qinmu Peng and
			Xinge You\textsuperscript{*},~\IEEEmembership{Senior Member,~IEEE,}
			\par
\thanks{\textsuperscript{*}Corresponding author: youxg@hust.edu.cn.}
\thanks{This work was supported in part by the National Key R\&D Program of China 2022YFC3301000, in part by the Fundamental Research Funds for the Central Universities, HUST: 2023JYCXJJ031.}
\thanks{X. Tang, S. Ye, Y. Shi, T. Hu, Q. Peng and X. You are with the School of Electronic Information and Communications, Huazhong University of Science and Technology, Wuhan 430074, China (E-mails: \{xiaolongtang, shuoye20, yufengshi17, huth, pengqinmu, youxg\}@hust.edu.cn).}}

% The paper headers
\markboth{IEEE TRANSACTIONS ON NEURAL NETWORKS AND LEARNING SYSTEMS}%
{Shell \MakeLowercase{\textit{et al.}}: A Sample Article Using IEEEtran.cls for IEEE Journals}

\maketitle

\begin{abstract}
Filter pruning has gained widespread adoption for the purpose of compressing and speeding up convolutional neural networks (CNNs). However, existing approaches are still far from practical applications due to biased filter selection and heavy computation cost. This paper introduces a new filter pruning method that selects filters in an interpretable, multi-perspective, and lightweight manner. Specifically, we evaluate the contributions of filters from both individual and overall perspectives. For the amount of information contained in each filter, a new metric called information capacity is proposed. Inspired by the information theory, we utilize the interpretable entropy to measure the information capacity, and develop a feature-guided approximation process. For correlations among filters, another metric called information independence is designed. Since the aforementioned metrics are evaluated in a simple but effective way, we can identify and prune the least important filters with less computation cost. We conduct comprehensive experiments on benchmark datasets employing various widely-used CNN architectures to evaluate the performance of our method. For instance, on ILSVRC-2012, our method outperforms state-of-the-art methods by reducing FLOPs by 77.4\% and parameters by 69.3\% for ResNet-50 with only a minor decrease in accuracy of 2.64\%.

\end{abstract}


\section{Introduction}
\IEEEPARstart{C}{onvolutional} neural networks (CNNs) have achieved impressive results in various computer vision tasks, including image classification \cite{krizhevsky2012imagenet,he2016deep, R2-Trans, ye2022discriminative, simonyan2014very, russakovsky2015imagenet, zhang2018shufflenet}, object detection \cite{girshick2015fast,redmon2016you, ren2015faster, liu2016ssd, zhu2021semantic}, action recognition \cite{yan2018spatial,li2019actional,tran2015learning,qiu2017learning,feichtenhofer2016convolutional} and automatic controll \cite{chen2013cooperative,chen2016terminal,zhu2021semantic}. However, their deployment on resource-constrained devices, such as embedded or mobile devices, is hindered by the high storage and computation cost. A mount of popular model compression techniques have been proposed to overcome these shortcomings, such as network pruning \cite{han2015learning,luo2017thinet,taylor,li2020group,filtersketch}, low-rank approximation\cite{zhang2015efficient,yin2021towards,lin2018holistic,hayashi2019exploring}, parameter quantization \cite{han2015learning,lin2020rotated,liu2020bi}, knowledge distillation \cite{tung2019similarity, park2019relational}. Among them, network pruning stands out as the most prevalent due to its remarkable ability to reduce model complexity.

Network pruning can be classified into two categories depending on the level of granularity: weight pruning and filter pruning. Weight pruning \cite{liu2018frequency, chen2018constraint} creates a sparse network by removing individual filter weights. Nevertheless, this method produces unstructured sparse patterns that result in discontinuous memory addressing, making it challenging to compress and accelerate the network on conventional hardware or with established sparse libraries like BLAS\cite{BLIS1}. In contrast, filter pruning \cite{he2017channel, yu2018nisp, sfp, guo2021gdp, liu2019metapruning} is well supported by existing hardware or software because it involves removing entire filters from the network. Therefore, filter pruning has garnered growing interest in recent studies.

Filter selection plays a pivotal role in the process of filter pruning. Initially, it is imperative to assess the significance of each filter, subsequently identifying and eliminating those deemed least crucial. An effective selection strategy is essential for retaining the network's predictive efficacy post-pruning. Nevertheless, existing methodologies for filter pruning encounter two noteworthy challenges in terms of filter selection:

\textit{1) Biased selection:} 
Current filter pruning approches can be classified into two categories based on different types of knowledge: intra-channel and inter-channel. Intra-channel methods directly evaluate the importance of each filter using knowledge from itself or the corresponding feature map, allowing for the removal of the least important filters. Li \textit{et al.} \cite {li2016pruning} calculate the significance of filters based on the ${l}_{1}$-norm, and filters with low-norm are considered unimportant. Lin \textit{et al.} \cite{lin2020hrank} compute the average rank of various feature maps produced by each filter and remove the filters with low-rank feature maps. In contrast, inter-channel methods calculate the correlation among different filters to measure the independence of each filter. Filters with weak independence are considered dispensable and can be eliminated. He \textit{et al.} \cite{he2019filter} measure the geometric median distance within a layer and remove filters that are in its proximity. Sui \textit{et al.} \cite{sui2021chip} evaluate the fungibility of every output feature map within a layer using the nuclear norm, and filters that exhibit high fungibility scores are identified as redundant. In contrast to inter-channel methodologies, intra-channel approaches may neglect the interdependencies among filters, potentially leading to the removal of filters containing unique information that is challenging to reacquire through alternative filters. In comparison, inter-channel methodologies effectively capture filter relationships to identify the most redundant filters. Nevertheless, these strategies face challenges in precisely assessing the informational content of individual filters, which may result in a substantial loss of information during the pruning process.

\textit{2) Heavy computation cost:} According to whether feature maps are required or not, existing methods can be categorized into two groups: data-free and data-driven. Data-free methods directly assess the significance of a filter by exclusively leveraging pre-trained weights, circumventing the need for feature maps. Li \textit{et al.} \cite{li2019exploiting} employ kernel sparsity and entropy, computed from filter weights, to evaluate each input channel. Molchanov \textit{et al.}\cite{taylor} approximate the contribution of a filter by computing first- and second-order Taylor expansions based on filter weights. Data-driven methods evaluate the significance of filters indirectly by utilizing the feature map generated by each filter. Hu \textit{et al.} \cite{hu2016network} measure the significance of the corresponding filter by calculating the proportion of zero activations in its corresponding feature map. Wang \textit{et al.}\cite{wang2018exploring} employ subspace clustering on feature maps to eliminate redundancy in convolutional filters. The feature map, which is generated by the filter and serves as an intermediate step in the neural network, reflects how the input data is transformed into the output labels by the network layer. Thus, it encapsulates information concerning the input data, output labels, and the characteristics of the filter itself \cite{lin2020hrank}. Consequently, the utilization of feature maps to gauge filter importance appears to offer greater effectiveness compared to data-free methodologies. However, data-driven approaches depend heavily on large datasets, requiring the computation of feature maps using extensive input data to reliably evaluate filter significance. This can lead to a significant increase in computational costs, especially with large datasets.

This paper introduces a novel filter pruning approach that considers multiple perspectives to capture the pertinent information from each filter and evaluates their significance in a computationally efficient manner. Our approach ensures that the pruned network preserves more essential information, which facilitates the recovery of the representational ability of the network during fine-tuning. The contributions of filters to the network are evaluated from both inter-channel and intra-channel perspectives. For each filter, the amount of information is measured by a metric named information capacity. Filters with higher information capacity are considered more informative and should be preserved. For interpretability and intuitiveness, we introduce information entropy as a measure of information capacity and experimentally demonstrate its validity. Furthermore, an effective approximation method is proposed to evaluate the entropy of feature maps using the corresponding filter weights, enabling the calculation of the filter's information capacity in a data-free manner. This allows for obtaining stable evaluation results without extensive computations from input data to feature maps. Additionally, we introduce a new metric called information independence to quantify the correlations among filters. Filters with greater information independence contain more unique information and should be preserved. By integrating these two metrics, we can identify and prune the least important filters. 

Compared with existing intra-channel methods \cite{li2016pruning, lin2020hrank} and inter-channel methods \cite{he2019filter, sui2021chip}, our method can evaluate the importance of each filter from multiple perspectives, resulting in more comprehensive evaluation outcomes. Experiments have demonstrated that our proposed approach exceeds existing data-free methods \cite{li2019exploiting, taylor} with respect to compression and acceleration, and even achieves superior results to the state-of-the-art data-driven approaches \cite{wang2018exploring, dubey2018coreset}. 

To sum up, our main contributions are as follows:
\begin{itemize}
  \item [1)] 
We analyze the limitations of current approaches from diverse angles, emphasizing the essential adherence to principles of ``maximum information" and ``independent information." Consequently, we introduce two metrics—information capacity and information independence—to appraise the significance of each filter, considering both individual and overarching perspectives.
  \item [2)]
During the process of filter estimation, we innovatively incorporate information entropy, propose a highly effective approximation, and empirically validate its efficacy. Consequently, we discerningly select filters within an interpretable and efficient framework.
  \item [3)]
Experiments on CIFAR-10/100 and ILSVRC-2012 substantiate the superior performance of our method in compressing diverse backbone networks, surpassing most state-of-the-art methods. For example, on ILSVRC-2012, our method outperforms state-of-the-art methods by reducing FLOPs by 77.4\% and parameters by 69.3\% for ResNet-50 with a marginal accuracy decrease of 2.64\%.
\end{itemize}

\begin{figure*}[htbp]
\centering
\includegraphics[width=1.0\textwidth]{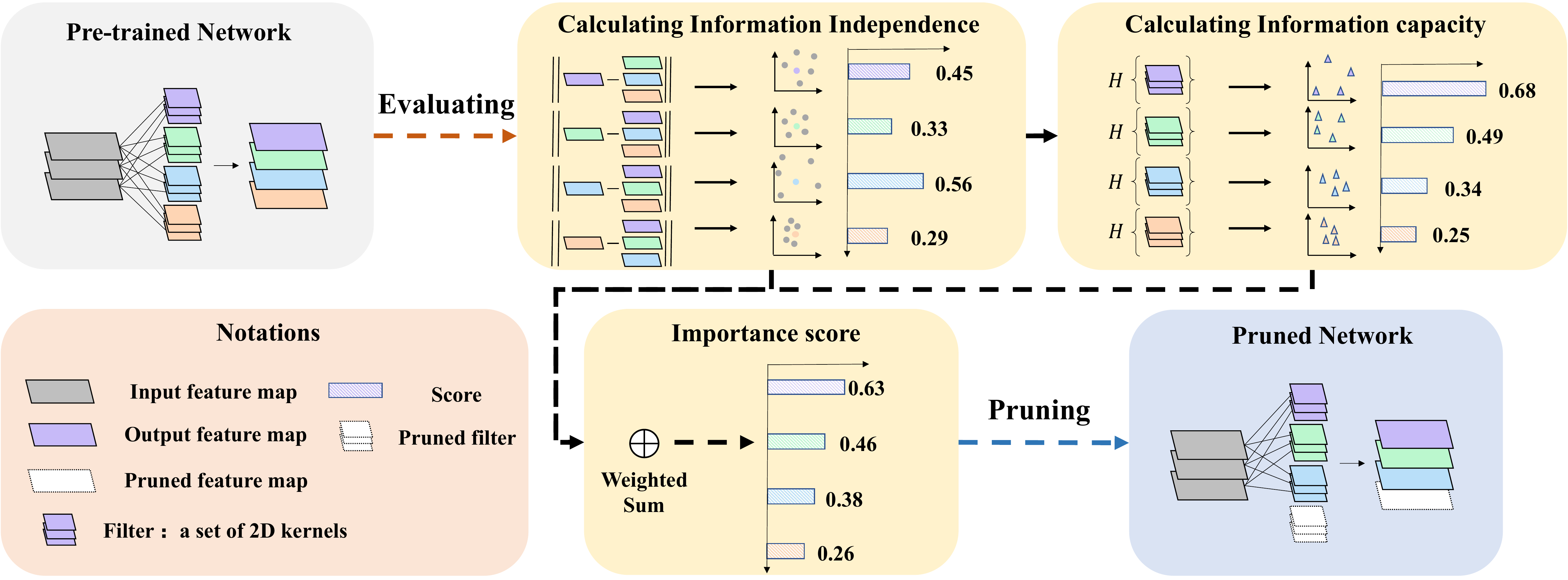}
\captionsetup{justification=raggedright}
\caption{Framework of the proposed approach. First, given a pre-trained network, we calculate the information capacity and information independence of each filter, respectively. Then, we weight and sum these two metrics to obtain an importance score for each filter. Finally, we prune the filters with the lowest scores according to a predetermined pruning rate, resulting in a compressed network.}
\label{fig_1}
\vspace{-0.4cm}
\end{figure*}

\section{Related Work}
\subsection{Intra-channel \& Inter-channel Filter Pruning}
Intra-channel methods directly calculate the importance of each filter using the knowledge from the filter itself or the corresponding feature map, and identify the least important filters for removal. He \textit{et al.} \cite{sfp} utilize the ${l}_{p}$-norm of different filters within each channel to assess their importance, allowing for potential updates in subsequent processes rather than immediate removal. Tang \textit{et al.} \cite{tang2020scop} evaluate the importance of each filter within a single channel by introducing knockoff features to compete with real features and comparing the results.

Unlike intra-channel methods, inter-channel methods determine filter importance using cross-channel information \cite{peng2019collaborative,tiwari2021chipnet,lemaire2019structured}. Peng \textit{et al.} \cite{peng2019collaborative} present a collaborative filter pruning algorithm that exploits inter-channel dependency information and constructs the pruning as a constrained 0-1 quadratic optimization problem. Nevertheless, these methods depend on the filters rather than the feature map data, potentially resulting in the insufficient identification and capture of critical feature characteristics. Tiwari \textit{et al.} \cite{tiwari2021chipnet} measure channel importance by regarding it as a constrained optimization problem and utilize nonlinear programming to solve it, while Lemaire \textit{et al.} \cite{lemaire2019structured} propose a knowledge distillation loss function using a barrier function formulation.

Diverging from conventional methodologies, we introduce two metrics, namely information capacity and information independence, designed to encapsulate intra-channel and inter-channel knowledge pertaining to filters, respectively. Consequently, our approach aims to retain maximal information within the pruned network, facilitating a more seamless recovery of accuracy during subsequent fine-tuning processes.

\vspace{-0.2cm}

\subsection{Data-driven \& Data-free Filter Pruning}
Certain studies \cite{zhuo2018scsp, ye2018rethinking, sfp}, characterized by their limited input datasets, fall into the category of data-free filter pruning. These studies endeavor to perform filter pruning based on the intrinsic characteristics of filters, such as feature mapping or geometric distribution. Zhuo \textit{et al.} \cite{zhuo2018scsp} determine whether the filter should be removed or retained based on spectral clustering. Ye \textit{et al.} \cite{ye2018rethinking} advocate the pruning of constant channels after random training, positing that smaller parameters or features bear lesser significance for the model. He \textit{et al.} \cite{sfp} make decisions regarding the preservation or removal of individual filters based on the ${l}_{2}$-norm criterion, and filters that are removed are considered for reintroduction or updates during subsequent training.

Some data-driven filter pruning works \cite{luo2017thinet,dubey2018coreset,suau2018principal} start from the feature map and attempt to extract richer knowledge to propose an effective importance evaluation. Luo \textit{et al.} \cite{luo2017thinet} retain a subset of filters in the next layer whose output can approximate the complete output, and remove other filters. Dubey \textit{et al.} \cite{dubey2018coreset} use quantization and Huffman coding to calculate the importance score, based on the absolute value of the activation weight post-training. Suau \textit{et al.} \cite{suau2018principal} utilize the eigenvectors and eigenvalues of the covariance matrix to calculate the energy factor, which can reflect the intrinsic correlation between filter responses and determine the importance of the corresponding filter. Compared to data-free methods, data-driven approaches typically achieve superior speed and accuracy concurrently. However, they require substantial input data for identifying filters to prune through feature maps, incurring a considerable cost during the pruning process.

Unlike these existing approaches, our method operates in a data-free manner but is guided by features. This is achieved through the novel introduction of information entropy, along with the proposal of a highly effective approximation.

\vspace{-0.2cm}

\subsection{Information Theory based Filter Pruning}
With the growing emphasis on interpretability in deep learning, information theory \cite{yu2020understanding, Yu2020Multivariate, shi2022deep, ye2023coping} is being applied to analyze the fundamental aspects of network models. As a result, some recent works on filter pruning aim to leverage information theory to guide their filter selection process. Min \textit{et al.}\cite{min20182pfpce} employ conditioned entropy, calculated as the entropy of feature maps conditioned on network loss, to prune filters. The higher the conditional entropy, the less important the filter, leading to the pruning of filters with high conditional entropy. Ganesh \textit{et al.} \cite{ganesh2021mint} measure the Mutual Information (MI) of filters between current layer and adjacent layer using conditional geometric mutual information based on graph criterion. Filters with high redundancy are pruned. Sarvani \textit{et al.} \cite{hrel} propose using mutual information between filters and class labels, denoted as relevance, calculated through Rényi's $\alpha$-order entropy, to quantify the significance of filters. Filters with elevated relevance are deemed more consequential. While approaches grounded in information theory are acknowledged for their intrinsic interpretability, their dependence on data-driven methodologies introduces a significant computational cost during the pruning process.

In contrast to prior studies, our approach is data-free yet feature-guided. We accomplish this by introducing information entropy and proposing a highly effective approximation for filter evaluation.

\begin{figure*}[htbp]
\centering
\subfloat[The difference of rank.]{
\includegraphics[width=0.48\textwidth]{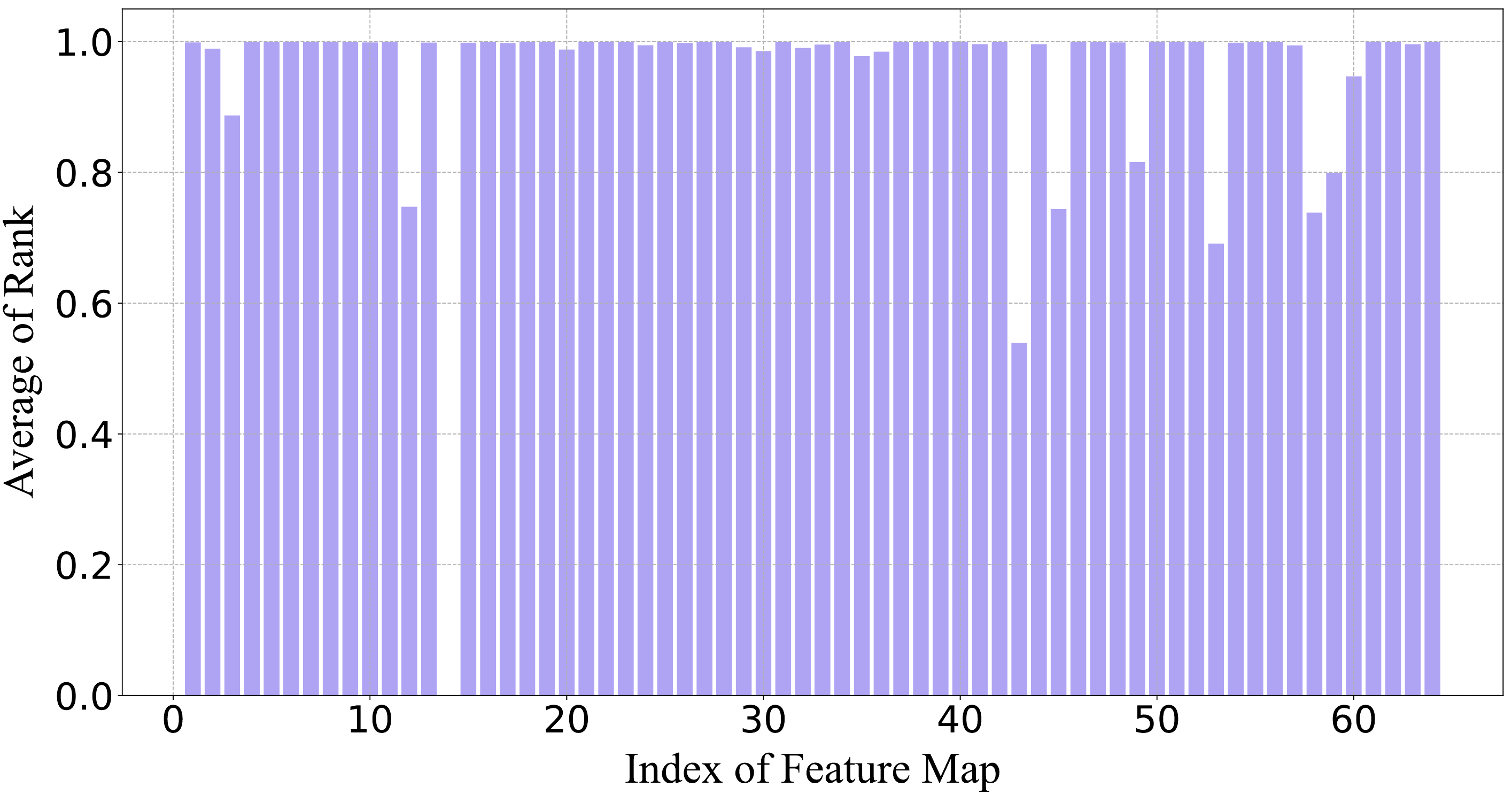}%
\label{fig_first_case}}
\quad
\subfloat[The difference of entropy.]{
\includegraphics[width=0.48\textwidth]{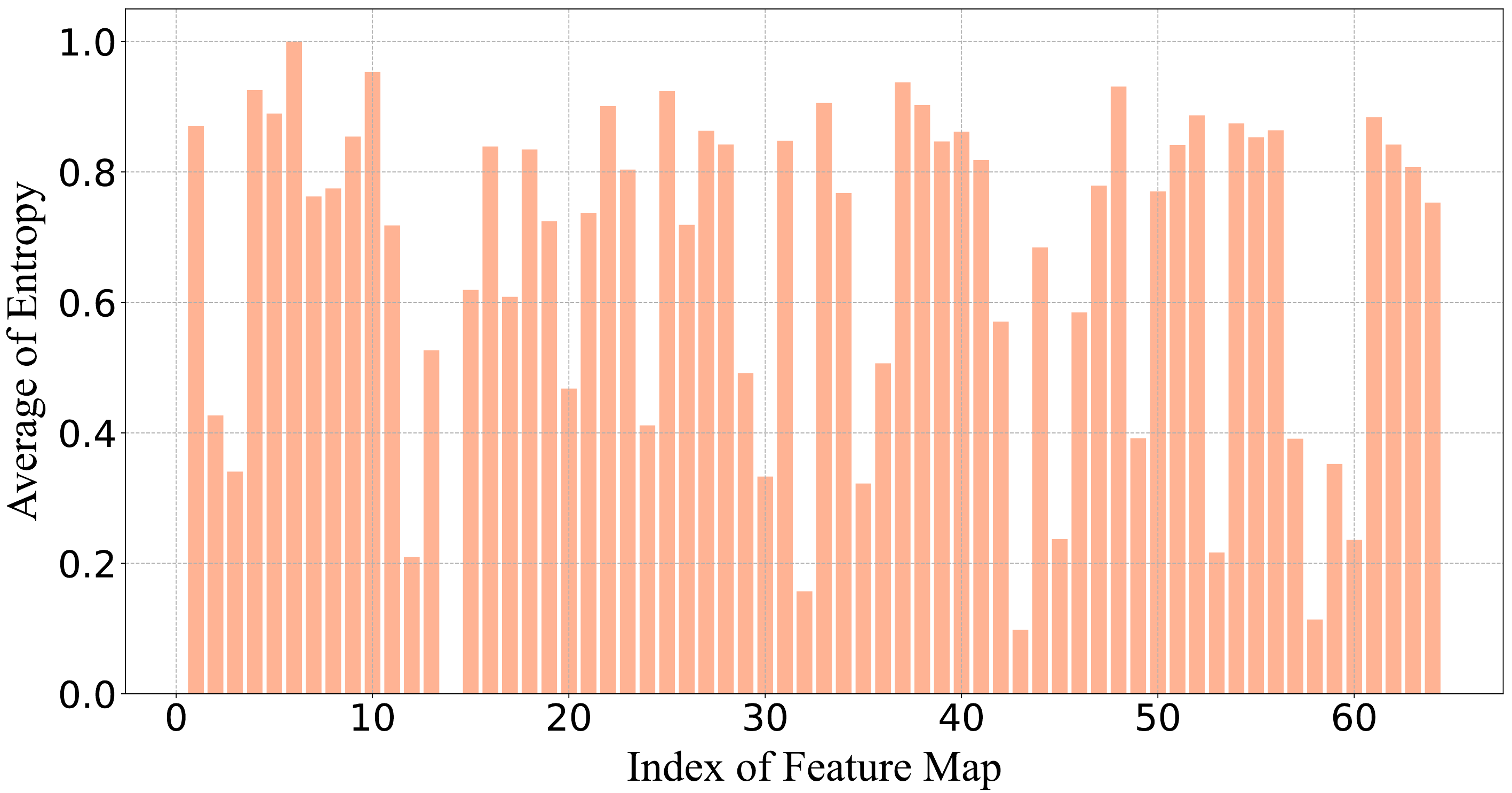}%
\label{fig_second_case}}
\captionsetup{justification=raggedright}
\caption{The graph displays the disparities in rank and entropy of feature maps produced from the same layer. The x-axis indicates the index of feature maps, and the y-axis represents the mean rank and entropy values of feature maps. The outcomes indicate that entropy can capture the knowledge contained in each feature map in a more detailed way, leading to better differentiation of the information conveyed by distinct feature maps.}
\label{fig_sim}
\vspace{-0.5cm}
\end{figure*}

\section{Proposed Approach}
\subsection{Notations}
Suppose there is a CNN model that is composed of $C$ convolutional layers, where the $j$-th convolutional layer has input and output channels of $n_{j - 1}$ and $n_{j}$, respectively. We use $F_{i,j} \in \mathbb{R}^{n_{j - 1} \times k_{j} \times k_{j}}$ to represent the $i$-th filter in the $j$-th convolutional layer, where $k_{j}$ is kernel size. Furthermore, $F_{i,j}$ can be represented as a set of 2D kernels $\left\{ w_{i,j}^{q},1 \leq q \leq n_{j - 1} \right\}$. We use $M_{i,j} = \left\{ m_{i,j}^{q},1 \leq q \leq s \right\} \in \mathbb{R}^{s \times h_{j} \times w_{j}}~$ represents the feature map generated by the filter $F_{i,j}$, where $s$ represents the number of input images, while $h_{j}$ and $w_{j}$ represent the height and width of the feature map, respectively. The $j$-th convolutional layer $L_{j}$ can be represented by $\left\{ F_{i,j},1 \leq i \leq n_{j} \right\} \in \mathbb{R}^{n_{j} \times n_{j - 1} \times k_{j} \times k_{j}}$. Then, $\left. \mathcal{L} = \big\{ \mathcal{L} \right._{j},1 \leq j \leq C \big\} \in \mathbb{R}^{{C \times n}_{j} \times n_{j - 1} \times k_{j} \times k_{j}}$ can be used to indicate the convolutional neural network.

Filter pruning aims to identify and remove insignificant filters from each layer of the network, thus obtaining a pruning network $\overset{-}{\mathcal{L}} = \big\{ \overset{-}{\mathcal{L}_{j}},1 \leq j \leq C \big\} \subseteq \mathcal{L}$. $\overset{-}{\mathcal{L}_{j}} = \big\{ F_{i,j},1 \leq i \leq \overset{-}{n_{j}} \big\} \subseteq \mathcal{L}_{j}$ represents the $j$-th convolutional layer . Assuming a pruning rate $p_{j}$ for the $j$-th layer, $\overset{-}{n_{j}} = \left\lceil {\left( 1 - p_{j} \right) \times n_{j}} \right\rceil$ denotes the number of remaining filters, where $\left\lceil \cdot \right\rceil$ stands for rounding up. Subsequently, the pruned network needs to undergo fine-tuning to restore its predictive power.

\vspace{-0.2cm}

\subsection{Algorithm}
The algorithm outlined in this paper is visually represented in Fig. 1. In the initial step, we compute the information capacity and information independence for each filter separately. Then, we integrate and assign weights to these two metrics to generate a holistic significance value for each filter. Ultimately, we remove the filters deemed least important, and the predictive performance of the pruned network is subsequently restored through the fine-tuning process.

\subsubsection{Information capacity}
We begin by computing the information capacity of the feature map. Some works measure the importance of feature maps by calculating their information richness. For example, \cite{lin2020hrank} uses the rank of a feature map to measure the amount of information within it and subsequently eliminate those filters with low-rank feature maps. Nonetheless, the rank of a matrix is merely an indication of the maximum number of linearly independent groups it contains and doesn't offer an intuitive measurement of the quantity of information present. Additionally, the calculation of rank adopts the $l_{0}$-norm as the singular value. This method explores the discriminative features of each feature map in a coarse-grained way, making it difficult to accurately reflect the differences between different feature maps. In other words, differences in the rank of feature maps are not sufficiently strong, which creates uncertainty in filter selection and limits the achievable acceleration and compression ratio.

We propose using entropy to measure the information capacity of feature maps, as it is a widely applicable and easily interpretable measure of information. For a system, entropy captures the linear and nonlinear relationships between variables, thus accurately reflecting the amount of information contained therein \cite{principe2010information}. Therefore, we argue that it is a better metric for measuring the information capacity. To demonstrate this, we compute the mean entropy and rank values of feature maps from the same layer. These feature maps are randomly selected from all feature maps in ResNet-50 trained on ILSVRC-2012. Specifically, we use the Rényi's $\alpha$-order entropy functional based on matrix \cite{Yu2020Multivariate} as a measure of entropy. As depicted in Fig. 2, the ranks differ little among each feature map, while the entropies have significant differences. Therefore, entropy captures the knowledge from each feature map in a fine-grained manner, and better distinguish the information carried by different feature maps.

However, it is challenging to directly substitute feature maps with filter weights for entropy calculation due to the complexity of high-dimensional variables' marginal probabilities. Therefore, we approximate the entropy of filters by quantifying the uncertainty encapsulated within the filter weights, which allows us to determine the corresponding information capacity. The essence of this approach lies in our recognition that entropy fundamentally reflects the degree of disorder or uncertainty in a system. The uncertainty of the filter can be characterized by the distribution of 2D kernels it contains. A more complex distribution of 2D kernels indicates a higher degree of uncertainty within the filter, enabling the extraction of more information and consequently yielding a feature map with greater information content. Therefore, the information capacity of the $i$-th filter in the $j$-th layer can be formulated as:
\begin{equation}
\setlength{\abovedisplayskip}{4pt}
\setlength{\belowdisplayskip}{1pt}
\label{ex1}
H_{f}\left( F_{i,j} \right) = H_{f}\left( {w_{i,j}^{1},\ldots,w_{i,j}^{n_{j - 1}}} \right)= - {\sum\limits_{q = 1}^{n_{j - 1}}{p_{q}{\log_{2}p_{q}}}} \text{,}
\end{equation}
\begin{equation}
\label{ex2}
p_{q} = \frac{exp\left( {sim}_{q} \right)}{\sum\limits_{q}^{n_{j - 1}}{exp\left( {sim}_{q} \right)}} \text{,}
\vspace{-0.2cm}
\end{equation}
where ${sim}_{q}$ represents the sum of Euclidean distance between the $q$-th kernel and other kernels in the filter, as delineated in Eq. (\ref{ex3}), wherein the Euclidean distance denotes the similarity between two kernels. 
\begin{equation}
\setlength{\abovedisplayskip}{4pt}
\setlength{\belowdisplayskip}{4pt}
\label{ex3}
{sim}_{q} = {\sum\limits_{{q}^{'}}^{n_{j - 1}}\left| \middle| w_{i,j}^{q} - w_{i,j}^{{q}^{'}} \middle| \right|_{2}} \text{.}
\end{equation}

Then we employ softmax function to calculate the probability distribution of all kernels. ${p}_{q}$ represents the probability of the q-th kernel. In order to distinguish from $q$ in Eq. (\ref{ex3}) and Eq. (\ref{ex4}), ${q}^{'}$ is also used to represent the index of the current kernel in the filter, with corresponding values ranging from $1$ to ${n_{j - 1}}$. The greater the probability, the smaller the similarity between the kernel and other kernels. To simplify the calculation, we can calculate the sum of the distances between each kernel and its nearest $M$ kernels to measure the distribution around this kernel in the filter. Hence, Eq. (\ref{ex4}) can be reformulated as:
\begin{equation}
\setlength{\abovedisplayskip}{4pt}
\setlength{\belowdisplayskip}{4pt}
\label{ex4}
{sim}_{q} \approx {\sum\limits_{{q}^{'}}^{M}\left| \middle| w_{i,j}^{q} - w_{i,j}^{{q}^{'}} \middle| \right|_{2}} \text{.}
\end{equation}

When the entropy of the filter, as defined above, is smaller,  it indicates that the corresponding feature map contains more information. Therefore, we can define the information capacity of the $i$-th filter in the $j$-th layer as:
\begin{equation}
\setlength{\abovedisplayskip}{5pt}
\setlength{\belowdisplayskip}{2pt}
\label{ex5}
{IM}_{cap}\left( F_{i,j} \right) = 1 - H_{f}\left( F_{i,j} \right) \text{.}
\end{equation}

We calculate the entropy of randomly selected feature maps and their corresponding filters from VGG-16 on CIFAR-10 and ResNet-50 on ILSVRC-2012, illustrating the relationship between them in Fig. 3. The plot reveals a strong correlation between the entropy of the feature map and that of the filter, as indicated by Pearson Correlation Coefficients of -0.90 and -0.83, respectively. Therefore, by employing Eq. (\ref{ex5}), we can systematically and efficiently compute the information capacity of various filters based on pre-trained weights, facilitating a ranking of the filters.

\subsubsection{Information independence}
To prevent the loss of special information that may be difficult to recover, we utilize information independence to assess the redundancy of information between the feature maps generated by filters. The information independence is computed using filter weights. When a filter is highly similar to its surrounding filters, the corresponding feature map also tends to be similar, indicating a stronger interchangeability of the contained information. We use the Euclidean distance as a metric for gauging the similarity between different filters. Thus, the information independence of the $i$-th filter in the $j$-th layer can be defined as follows:
\begin{equation}
\setlength{\abovedisplayskip}{5pt}
\setlength{\belowdisplayskip}{4pt}
\label{ex6}
{IM}_{ind}\left( F_{i,j} \right) = {\sum\limits_{{i}^{'}}^{n_{j}}\left| \middle| F_{i,j} - F_{{i}^{'},j} \middle| \right|_{2}} \text{,}
\end{equation}
where ${i}^{'}$  is also used to represent the index of the filter in the $j$-th layer, solely for the purpose of differentiation from $i$.

\begin{figure}[!t]
\centering
\includegraphics[width=.49\textwidth]{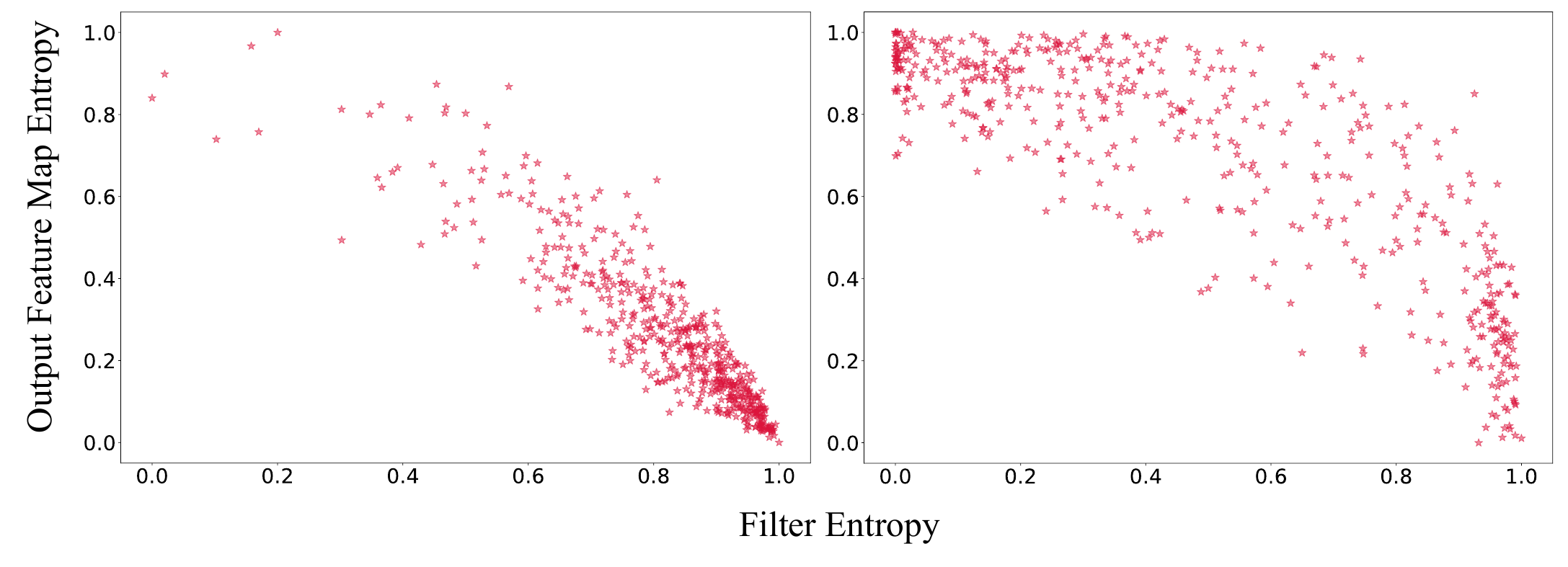}
\captionsetup{justification=raggedright}
\caption{The correlation between feature map entropy and corresponding filter entropy. The left diagram pertains to VGG-16 on CIFAR-10, while the right diagram pertains to ResNet-50 on ILSVRC-2012.}
\label{fig_3}
\vspace{-0.63cm}
\end{figure}

\subsubsection{Integration}
Finally, we use information capacity and information independence together to measure the significance of each filter. Therefore, the importance score of the $i$-th filter in the $j$-th layer can be formulated as:
\begin{multline}
\setlength{\abovedisplayskip}{8pt}
\setlength{\belowdisplayskip}{8pt}
\label{ex7}
\mathcal{O}\left( F_{i,j} \right) = \sigma \cdot Norm \Big( {IM}_{cap}\left( F_{i,j} \right)\Big) \\
+ (1 - \sigma) \cdot Norm \Big( {IM}_{ind}\left( F_{i,j} \right) \Big) \text{,}
\end{multline}
where $\sigma$ is a weight index used to control the effects of information capacity and information independence on the importance score of the filter. $Norm\left(\cdot \right)$ stands for min-max normalization. Please refer to Section. \uppercase\expandafter{\romannumeral 4} for detailed discussion and experiments on $\sigma$.

\subsection{Pruning Strategy}
Given a pruning rate ${p}_{j}$ for each layer, we retain the most important $\overset{-}{n_{j}}$ filters while removing the rest. We perform pruning for all layers at once, and the pruned filters are removed from the network structure. It should be noted that each neuron of the fully connected layer can be considered as a filter with a kernel size of $1\times1$. Therefore, the fully connected layer can be pruned in a similar way as the convolutional layer. Algorithm 1 depicts the complete framework of our algorithm.

\begin{algorithm}[H]
\caption{Filter Pruning based on Information Capacity and Independence.}\label{alg:alg1}
\begin{algorithmic}
\STATE 
\STATE \hspace{0.0cm} \textbf{Input: } A pre-trained $C$-layer model $L$ with filter set $F$, a set of pruning rates $p = \left\{ p_{1},p_{1},\ldots,p_{C} \right\}. $
\STATE \hspace{0.cm} \textbf{Output: } A compressed network $\overset{-}{\mathcal{L}}. $
\STATE \hspace{0.cm} \textbf{for} $\left. j = 1\rightarrow C \right.$ \textbf{do}
\STATE \hspace{0.5cm} \textbf{for} $\left. i = 1\rightarrow n_{j} \right.$ \textbf{do}
\STATE \hspace{1.0cm} Calculate the filter importance $\mathcal{O}\left( F_{i,j} \right)$ via Eq. (\ref{ex7});
\STATE \hspace{0.5cm} \textbf{end for}
\STATE \hspace{0.0cm} \textbf{end for}
\STATE \hspace{0.cm} \textbf{for} $\left. j = 1\rightarrow C \right.$ \textbf{do}
\STATE \hspace{0.5cm} $\overset{-}{n_{j}} = \left\lceil {\left( 1 - p_{j} \right) \times n_{j}} \right\rceil $;
\STATE \hspace{0.5cm} Remove $\left( n_{j} - \overset{-}{n_{j}} \right)$ filters with the smallest importance score in $F = \left\{ F_{1,j},F_{2,j},\ldots,F_{n_{j},j} \right\}$;
\STATE \hspace{0.0cm} \textbf{end for}
\STATE \hspace{0.0cm} \textbf{Return: } The compressed network $\overset{-}{\mathcal{L}}$ after fine-pruning.
\end{algorithmic}
\label{alg1}
\end{algorithm}

\section{Experiments}
We conduct multiple sets of experiments on different benchmark image classification datasets, including comparative experiments and ablation experiments, to demonstrate the effectiveness and rationality of our method. Specifically, we conducted pruning experiments on three representative networks, including VGG-16 \cite{simonyan2014very} and ResNet-56/110 \cite{he2016deep} on CIFAR-10 \cite{krizhevsky2009learning}, as well as VGG-16 on CIFAR-100 \cite{krizhevsky2009learning}. Furthermore, we utilize our approach to prune the ResNet-50 \cite{he2016deep} architecture and evaluate its efficacy against state-of-the-art pruning approaches proposed in recent years, on the ImageNet ILSVRC-2012 dataset \cite{russakovsky2015imagenet}. This dataset serves as a benchmark for large-scale image classification tasks. Additionally, we apply the proposed method to transformer-like models and instance segmentation tasks, with corresponding datasets being ILSVRC-2012 and MS COCO\cite{mscoco}, respectively.

\begin{table*}[htbp]
\caption{Results of VGG-16 and ResNet-56/110 on CIFAR-10. Dashed lines depict various pruning rates applied to the same dataset and network.\label{tab:table1}}
\centering
\setlength{\tabcolsep}{5.0mm}{
\begin{tabular}{ccccccc}
\toprule[1pt]
\multirow{2}{*}[-0.6ex]{Model} & \multirow{2}{*}[-0.6ex]{Method} & \multicolumn{3}{c}{Top-1 Accuracy(\%)} & \multirow{2}{*}[-0.6ex]{FLOPs[M] / PR[\%]} & \multirow{2}{*}[-0.6ex]{Params[M] / PR[\%]} \\
\cmidrule{3-5}
& & Baseline & Pruned & \( \Delta \) \\
\midrule 
\multirow{16}{*}{VGG-16} 
& L1\cite {li2016pruning} & 93.25 & 93.40 & 0.15 & 206.00 / 34.3 & 5.40 / 64.0 \\
& SSS\cite{huang2018data} & 93.96 & 93.02 & -0.97 & 183.13 / 41.6 & 3.93 / 73.8 \\
& GAL-0.1 \cite{lin2019towards} &93.96 &93.42 &-0.54 &171.89 / 45.2 &2.67 / 82.2 \\
& Hrank\cite{lin2020hrank} &93.96 &93.43&-0.53&145.61 / 53.5 &2.51 / 82.9 \\
& CHIP\cite{sui2021chip} &93.96 &93.86 &-0.10 &131.17 / 58.1 &2.76 / 81.6 \\
& Switch\cite{switch} & 93.58 & 93.76 & 0.18 & 206.00 / 34.3 & 5.40 / 64.0 \\
& \textbf{Ours} &\textbf{93.96} &\textbf{94.30} &\textbf{0.34} &\textbf{129.23 / 58.9} &\textbf{2.53 / 83.1} \\ 
\cdashline{2-7}[3pt/2pt] 
& Hrank\cite{lin2020hrank} &93.96 &92.34 &-1.62 &108.61 / 65.3 &2.64 / 82.1 \\
& CHIP\cite{sui2021chip} &93.96 &93.72 &-0.24 &104.78 / 66.6 &2.50 / 83.3 \\
& Corse\cite{Coarse-to-Fine} &93.96 &93.79 &-0.17 &94.15 / 70.1 &1.79 / 88.1 \\
& \textbf{Ours} &\textbf{93.96} &\textbf{94.02} &\textbf{0.06} &\textbf{102.87 / 67.3} &\textbf{2.27 / 84.9} \\
\cdashline{2-7}[3pt/2pt] 
& Hrank\cite{lin2020hrank} &93.96 &91.23 &-2.73 &73.70 / 76.5 &1.78 / 92.0 \\
& HRel\cite{hrel} &93.90 &93.40 &-0.50 &47.51 / 84.9 &0.75 / 95.0 \\
& CHIP\cite{sui2021chip} &93.96 &93.18 &-0.78 &66.98 / 78.6 &1.90 / 87.3 \\
& FSIM-E\cite{FSIM-E} &93.96 &92.84 &-1.12 &59.90 / 80.9 &2.32 / 84.2 \\
& RGP\cite{rgp} &93.14 &92.76 &-0.38 &78.42 / 75.0 &3.81 / 74.9 \\
& \textbf{Ours} &\textbf{93.96} &\textbf{93.67} &\textbf{-0.29} &\textbf{65.94 / 79.0} &\textbf{1.77 / 88.2} \\
\midrule
\multirow{16}{*}{ResNet-56} 
& L1\cite {li2016pruning} &93.26 &93.06 &0.20 &90.90 / 27.6 &0.73 / 14.1 \\
& NISP\cite{yu2018nisp} &93.04 &93.01 &-0.03 &81.00 / 35.5 &0.49 / 42.4 \\
& GAL-0.6\cite{lin2019towards} &93.26 &93.38 &0.12 &78.30 / 37.6 &0.75 / 11.8 \\
& Hrank\cite{lin2020hrank} &93.26 &93.17 &-0.09 &62.72 / 50.0 &0.49 / 42.4 \\
& FilterSketch\cite{filtersketch} &93.26 &93.19 &-0.07 &73.36 / 41.5 &0.50 / 41.2 \\
& CLR-RNF-0.56\cite{CLRRNF} &93.26 &93.27 &0.01 &54.00 / 57.3 &0.38 / 55.5 \\
& White-Box\cite{white-box} &93.26 &93.54 &0.28 &56.69 / 55.6 & $-$ \\
& Switch\cite{switch} &93.59 &93.56 &-0.03 &59.40 / 52.6 & $-$ \\
& \textbf{Ours} &\textbf{93.26} &\textbf{93.53} &\textbf{0.27} &\textbf{52.84 / 58.6} &\textbf{0.36 / 57.6} \\
\cdashline{2-7}[3pt/2pt] 
& GAL-0.8\cite{lin2019towards} &93.26 &91.58 &-1.68 &49.99 / 60.2 &0.29 / 65.9 \\
& Hrank\cite{lin2020hrank} &93.26 &90.72 &-2.54 &32.50 / 74.1 &0.27 / 68.1 \\
& HRel\cite{hrel} &93.80 &93.19 &-0.61 &47.57 / 62.1 &0.30 / 63.8 \\
& CHIP\cite{sui2021chip} &93.26 &92.05 &-1.21 &34.79 / 72.3 &0.24 / 71.8 \\
& FSIM-E\cite{FSIM-E} &93.30 &91.96 &-1.34 &31.40 / 75.4 &0.22 / 74.1 \\
& RGP\cite{rgp} &93.26 &91.50 &-1.76 &31.54 / 75.3 &0.21 / 74.5 \\
& \textbf{Ours} &\textbf{93.26} &\textbf{92.49} &\textbf{-0.77} &\textbf{34.09 / 73.3} &\textbf{0.22 / 74.1} \\
\midrule 
\multirow{10}{*}{ResNet-110} 
& L1\cite {li2016pruning} &93.50 &93.30 &-0.20 &155.00 / 38.7 &1.16 / 32.6 \\
& GAL-0.5\cite{lin2019towards} &93.50 &93.74 &0.24 &130.20 / 48.5 &0.95 / 44.8 \\
& Hrank\cite{lin2020hrank} &93.50 &92.65 &-0.85 &79.30 / 68.6 &0.53 / 68.7 \\
& HRel\cite{hrel} &93.50 &93.03 &-0.47 &95.72 / 62.1 &0.62 / 63.8 \\
& FilterSketch\cite{filtersketch} &93.57 &93.44 &-0.13 &92.84 / 63.3 &0.69 / 59.9 \\
& CHIP\cite{sui2021chip} &93.50 &93.63 &0.13 &71.69 / 71.6 &0.54 / 68.3 \\
& CLR-RNF-0.69\cite{CLRRNF} &93.57 &93.71 &0.14 &86.80 / 66.0 &0.53 / 69.1 \\
& FSIM-E\cite{FSIM-E} &93.58 &93.68 &0.10 &73.59 / 71.1 &0.58 / 66.5 \\
& Switch\cite{switch} &93.68 &93.38 &-0.30 &95.64 / 62.8 &$-$ \\
& \textbf{Ours} &\textbf{93.50} &\textbf{93.78} &\textbf{0.28} &\textbf{72.24 / 71.9} &\textbf{0.53 / 69.4} \\
\bottomrule[1pt] 
\end{tabular}}
\vspace{-0.25cm}
\end{table*}

\vspace{-0.15cm}
\subsection{Implementation Details}
\textit{1) Training Strategy:} 
We use PyTorch \cite{paszke2017automatic} 1.12 framework to implement our method. We train the models on NVIDIA GeForce RTX 2080Ti GPUs with 11GB of RAM for fine-tuning. After determining which filters need to be pruned and performing the pruning, we utilize the Stochastic Gradient Descent (SGD) algorithm as our gradient descent strategy to expedite convergence. For the CIFAR-10/100 dataset, we perform network retraining for 300 epochs, with configuration parameters such as batch size, initial learning rate, weight decay, and momentum  set to 128, 0.01, 0.05, and 0.9, respectively. Concurrently, on the ImageNet ILSVRC-2012 dataset, we execute network retraining for 180 epochs with configuration parameters set as follows: 256 for batch size, 0.1 for initial learning rate, 0.0001 for weight decay, and 0.99 for momentum. For a comprehensive understanding of the experimental setup related to transformer-like models and instance segmentation tasks, please consult the respective sections, namely Subsections D and E.

\textit{2) Performance Metric:}
Aligned with earlier research methodologies, we assess model complexity through the metrics of floating-point operations (FLOPs) and model parameters. The pruning ratio (PR) of these metrics signifies the compression rate post-pruning, while \( \Delta \) denotes the accuracy reduction subsequent to pruning. We evaluate the performance of both the original and pruned networks across CIFAR-10/100 and ImageNet ILSVRC-2012, employing the Top-1 accuracy metric. For ImageNet, we also consider the Top-5 accuracy. Furthermore, on the MS COCO dataset, performance is scrutinized using the box mAP and mask mAP metrics.

\subsection{Results on CIFAR-10}
We employ our proposed method to evaluate its effectiveness on commonly utilized CNNs for CIFAR-10, specifically targeting VGG-16 and ResNet-56/110 architectures. It is crucial to note that the particular configuration of VGG-16 employed in our experiments is explicitly outlined in \cite{li2016pruning}.

\begin{table*}[!t]
\caption{Results of ResNet-50 on ILSVRC-2012. Dashed lines depict various pruning rates applied to the same dataset and network.\label{tab:table2}}
\centering
\setlength{\tabcolsep}{3.1mm}{
\begin{tabular}{ccccccccc}
\toprule[1pt]
\multirow{2}{*}[-0.6ex]{Method} & \multicolumn{3}{c}{Top-1 Accuracy(\%)} & \multicolumn{3}{c}{Top-5 Accuracy(\%)} & \multirow{2}{*}[-0.6ex]{FLOPs[G] / PR[\%]} & \multirow{2}{*}[-0.6ex]{Params[M] / PR[\%]} \\
\cmidrule{2-7}
& Baseline & Pruned & \( \Delta \) & Baseline & Pruned & \( \Delta \) \\
\midrule
SSS-32\cite{huang2018data} &76.13 &74.18 &-1.95 &92.86 &91.91 &-0.95 &2.82 / 31.1 &18.60 / 27.1  \\
GAL-0.5\cite{lin2019towards} &76.13 &71.95 &-4.18 &92.86 &90.94 &-1.92 &2.33 / 43.0 & 21.20 / 16.9 \\
Hrank\cite{lin2020hrank} &76.15 &74.98 &-1.17 &92.87 &92.33 &-0.54 &2.30 / 43.8 & 16.15 / 36.7 \\
HRel\cite{hrel} &76.15 &75.47 &-0.68 &92.87 &92.60 &-0.27 &2.11 / 48.7 & 13.23 / 48.2 \\
FilterSketch-0.6\cite{filtersketch} &76.13 &74.68 &-1.45 &92.86 &92.17 &-0.69 &2.23 / 45.5 &14.53 / 43.0 \\
CLR-RNF-0.20\cite{CLRRNF} &76.01 &74.85 &-1.16 &92.96 &92.31 &-0.65 &2.45 / 40.4 & 16.92 / 33.8 \\
White-Box\cite{white-box} &76.15 &75.32 &-0.83 &92.96 &92.43 &-0.53 &2.22 / 45.6 & $-$ \\
Corse\cite{Coarse-to-Fine} &76.18 &76.23 &0.05 &92.79 &92.87 &0.08 &2.48 / 40.0 & 21.56 / 15.6 \\
Switch\cite{switch} &76.15 &75.67 &-0.48 &92.87 &92.43 &-0.06 &2.38 / 42.2 & $-$ \\
RGP\cite{rgp} &76.22 &75.30 &-0.92 &$-$ &$-$ &$-$ &2.32 / 43.8 & 14.38 / 43.8 \\
\textbf{Ours} &\textbf{76.13} &\textbf{76.08} &\textbf{-0.05} &\textbf{92.86} &\textbf{92.85} &\textbf{-0.01}  &\textbf{2.05 / 50.4}  &\textbf{13.83 / 45.9}\\
\hdashline[3pt/2pt] 
GAL-0.5-joint\cite{lin2019towards} &76.13 &71.80 &-4.33 &92.86 &90.82 &-2.04 &1.84 / 55.0 &19.31 / 24.3 \\
Hrank\cite{lin2020hrank} &76.15 &71.98 &-4.17 &92.87 &91.01 &-1.86 &1.55 / 62.3 &13.37 / 47.7 \\
HRel\cite{hrel} &76.15 &73.67 &-2.48 &92.87 &92.12 &-0.75 &1.38 / 66.4 & 9.10 / 64.4 \\
FilterSketch-0.4\cite{filtersketch} &76.13 &73.04 &-3.09 &92.86 &91.18 &-1.68 &1.51 / 63.1 &10.40 / 59.2 \\
CLR-RNF-0.44\cite{CLRRNF} &76.01 &72.67 &-3.34 &92.96 &91.09 &-1.87 &1.23 / 70.1 & 9.00 / 64.8 \\

White-Box\cite{white-box} &76.15 &74.21 &-1.94 &92.96 &92.01 &-0.95 &1.50 / 63.8 & $-$ \\
FSIM-E\cite{FSIM-E} &76.15 &75.32 &-0.83 &92.87 &92.45 &-0.42 &1.75 / 57.2 & 11.98 / 53.1 \\
Switch\cite{switch} &76.15 &74.86 &-1.29 &92.87 &92.43 &-0.44 &1.92 / 53.5 & $-$ \\
RGP\cite{rgp} &76.22 &74.07 &-2.15 &$-$ &$-$ &$-$ &1.55 / 62.5 & 9.59 / 62.5 \\

\textbf{Ours} &\textbf{76.13} &\textbf{75.01} &\textbf{-1.12} &\textbf{92.86} &\textbf{92.30} &\textbf{-0.56}  &\textbf{1.50 / 63.8}  &\textbf{10.81 / 57.7}\\
\hdashline[3pt/2pt] 
GAL-1-joint\cite{lin2019towards} &76.13 &69.31 &-6.82 &92.86 &89.12 &-3.74 &1.11 / 72.9 &10.21 / 60.0 \\
Hrank\cite{lin2020hrank} &76.15 &69.10 &-7.05 &92.87 &89.58 &-3.29 &0.98 / 76.0 &8.27 / 67.6 \\
FilterSketch-0.2\cite{filtersketch} &76.13 &69.43 &-6.70 &92.86 &89.23 &-3.63 &0.93 / 77.3 &7.18 / 71.8 \\
CLR-RNF-0.52\cite{CLRRNF} &76.01 &71.11 &-4.90 &92.96 &90.42 &-2.54 &0.93 / 77.4 & 6.90 / 73.0 \\
RGP\cite{rgp} &76.22 &72.68 &-3.54 &$-$ &$-$ &$-$ &1.03 / 75.0 & 6.39 / 75.0 \\
\textbf{Ours} &\textbf{76.13} &\textbf{73.49} &\textbf{-2.64} &\textbf{92.86} &\textbf{91.45} &\textbf{-1.41}  &\textbf{0.93 / 77.4}  &\textbf{7.85 / 69.3}\\
\bottomrule[1pt] 
\end{tabular}}
\vspace{-0.3cm}
\end{table*}

\textit{1) VGG-16:}
As shown in Table \uppercase\expandafter{\romannumeral1}, our method is compared against several state-of-the-art methods\cite{li2016pruning, huang2018data,lin2019towards, lin2020hrank, sui2021chip}, and surpasses them in terms of both retaining accuracy and reducing model complexity. Notably, our method achieves a 0.34\% improvement in accuracy compared to the original baseline model, while reducing FLOPs by 58.9\% and parameters by 83.1\%. Despite applying two more aggressive compression levels resulting in 67.3\% and 79.0\% FLOPs reduction and 84.9\% and 88.2\% parameter reduction, our method still achieves outstanding performance. In fact, our method achieves accuracy gains of 1.68\% and 2.44\%, respectively, over HRank \cite{lin2020hrank}, and 0.30\% and 0.49\%, respectively, over CHIP \cite{sui2021chip}, with comparable compression rates maintained.

\textit{2) ResNet-56:}
Table \uppercase\expandafter{\romannumeral1} depicts the pruning results for ResNet-56. We begin by applying our method to ResNet-56, and the effectiveness is demonstrated by achieving an accuracy of 93.53\% while reducing FLOPs and parameters by 58.0\% and 57.6\% respectively. Notably, our method achieves significantly better model complexity reduction than L1 \cite{li2016pruning} (58.0\% vs. 27.6\% for FLOPs and 57.6\% vs. 14.1\% for parameters) while both obtaining nearly the same accuracy increase. These results suggest that our approach can effectively reduce the amount of parameters and computational cost of the model while maintaining its performance as much as possible. Furthermore, our approach exhibits higher accuracy of 93.53\% when compared to CLR-RNF-0.69 \cite{CLRRNF}, with a similar compression ratio. In contrast, CLR-RNF-0.69 achieved an accuracy of 93.27\%. This result clearly demonstrates the effectiveness of our method.

\textit{3) ResNet-110:}
Table \uppercase\expandafter{\romannumeral1} depicts the results of our pruning experiments on ResNet-110. Similar to ResNet-56, our approach enhances the accuracy of the baseline model to 93.78\% from 93.50\%, while simultaneously reducing FLOPs by approximately 71.9\% and parameters by 69.4\%. Our method significantly outperforms GAL-0.5 \cite{lin2019towards} in terms of model complexity reduction, achieving reductions of 71.9\% in FLOPs and 69.4\% in parameters, surpassing GAL-0.5's reductions of 48.5\% and 44.8\%, respectively. Moreover, our approach results in a slight increase in accuracy of 0.04\%. Furthermore, our method outperforms CHIP \cite{sui2021chip} with respect to accuracy performance of 93.78\% despite CHIP achieving slightly better complexity reduction. For instance, CHIP achieved an accuracy of 93.63\%. This superior performance is attributable to our method's comprehensive evaluation of filter importance, which considers multiple aspects within the pre-trained weights, including both information capacity and information independence.

\subsection{Results on CIFAR-100}
Table \uppercase\expandafter{\romannumeral3} presents the pruning results for VGG-16 on CIFAR-100. According to the findings, our approach outperforms the state-of-the-art method \cite{rgp} by excelling in both accuracy retention and model complexity reduction. Significantly, through careful reduction of model complexity, our method incurs only a minimal 0.05\% Top-1 accuracy loss, concurrently achieving substantial reductions of 67.3\% in FLOPs and 84.6\% in parameters. With further reduction in model complexity, our method achieves an optimal balance in accuracy and compression ratio. For instance, although our compressed model shows a slightly greater loss in accuracy than RGP (-2.52\% vs. -2.50\%), we achieve superior compression ratios in terms of FLOPs (84.8\% vs. 84.3\%) and parameters (89.2\% vs. 84.4\%), and also maintain a higher accuracy.

\begin{table}[htbp]
\caption{Results of VGG-16 on CIFAR-100. Dashed lines depict various pruning rates applied to the same dataset and network. \label{tab:table1}}
\centering
\setlength{\tabcolsep}{0.7mm}{
\begin{tabular}{cccccc}
\toprule[1pt]
\multirow{2}{*}[-0.6ex]{Method} & \multicolumn{3}{c}{Top-1 Accuracy(\%)} & \multirow{2}{*}[-0.6ex]{FLOPs[M] / PR[\%]} & \multirow{2}{*}[-0.6ex]{Params[M] / PR[\%]} \\
\cmidrule{2-4}
& Baseline & Pruned & \( \Delta \) \\
\midrule 
RGP\cite{rgp} &71.83 &71.15 &-0.68 &98.65 / 68.7 &4.78 / 68.7 \\
\textbf{Ours} &\textbf{73.06} &\textbf{73.04} &\textbf{-0.05} &\textbf{102.88 / 67.3} &\textbf{2.31 / 84.6} \\ 
\hdashline[3pt/2pt] 
RGP\cite{rgp} &71.83 &70.40 &-1.43 &79.00 / 74.9 &3.82 / 75.0 \\
\textbf{Ours} &\textbf{73.06} &\textbf{71.62} &\textbf{-1.44} &\textbf{65.76 / 79.1} &\textbf{1.81 / 88.0} \\
\hdashline[3pt/2pt] 
RGP\cite{rgp} &71.83 &69.33 &-2.50 &49.55 / 84.3 &2.39 / 84.4 \\
\textbf{Ours} &\textbf{73.06} &\textbf{70.54} &\textbf{-2.52} &\textbf{47.82 / 84.8} &\textbf{1.63 / 89.2} \\
\bottomrule[1pt] 
\end{tabular}}
\vspace{-0.5cm}
\end{table}

\subsection{Results on ILSVRC-2012}
\vspace{-0.02cm}
\textit{1) ResNet-50:}
On the ILSVRC-2012 dataset, both our method and other recent methods are applied to ResNet-50, and the results are shown in Table \uppercase\expandafter{\romannumeral2}. According to the results, our method achieves a significant reduction in model complexity with only a minor trade-off in accuracy. Specifically, when appropriately reducing model complexity, our method incurs a mere 0.05\% Top-1 accuracy loss and 0.01\% Top-5 accuracy loss, while achieving substantial reductions of 50.4\% FLOPs and 45.9\% parameters. Even with further reduction in model complexity, our method still outperform other state-of-the-art approaches. For instance, our method is 0.54$×$ the Top-1 accuracy drop of CLR-RNF-0.52 (-2.64 vs -4.90), and 0.56$×$ the Top-5 accuracy drop of CLR-RNF-0.52 (-1.41 vs -2.54), while enjoying similar model size reduction of 69.3\% and FLOPs reduction of 77.4\%. Table \uppercase\expandafter{\romannumeral4} presents the decrease in inference time. Our approach demonstrates notable improvements in inference speed without compromising overall performance significantly. To illustrate, it attains a 1.89× acceleration in GPU processing speed with only a 2.59\% reduction in Top-1 accuracy when compared to the baseline. Consequently, our approach demonstrates excellent performance on even the most intricate datasets.
\vspace{-0.1cm}

\begin{table}[htbp]
\caption{The FLOPS, GPU time, Top-1 accuracy and acceleration ratio of our ResNet-50 pruning approach. The corresponding network runs on a solitary NVIDIA RTX 2080Ti GPU with a batch size of 32.\label{tab:table4}}
\centering
\setlength{\tabcolsep}{1.8mm}{
\begin{tabular}{ccccc}
\toprule[1pt]
\multirow{1}{*}[-0.6ex]{Model} & \multirow{1}{*}[-0.6ex]{FLOPs[G]} & \multirow{1}{*}[-0.6ex]{Top-1(\%)} & \multirow{1}{*}[-0.6ex]{GPU time(ms)} & \multirow{1}{*}[-0.6ex]{Acceleration} \\
\midrule 
ResNet-50 \cite{he2016deep} & 4.13 & 76.13 & 8.92 & 0.00× \\
Ours & 2.05 & 76.08 & 6.41 & 1.39× \\
Ours & 1.50 & 75.01 & 5.72 & 1.56× \\
Ours & 0.93 & 73.49 & 4.72 & 1.89× \\
\bottomrule[1pt] 
\end{tabular}}
\vspace{-0.05cm}
\end{table}

\textit{2) Transformer-based model:}
In recent years, vision transformers, exemplified by models such as ViT \cite{vit}, Swin-Transformer \cite{swin}, DeiT \cite{deit} and LVVIT \cite{LVVIT}, have prominently emerged in various visual tasks, including image classification and object detection. Among them, DeiT and LVVIT have achieved outstanding classification accuracy with relatively lower data and parameter requirements, indicating their notable value in practical applications. In our endeavor to further assess the effectiveness of our approach, we conducted pruning on DeiT and LVVIT using the ILSVRC-2012 dataset. Inspired by \cite{dynamic}, which proposes a highly effective strategy for shifting the focus of pruning methods from CNNs to Transformers, we employ Token Sparsity. Specifically, we determine the importance of each token based on its information capacity and independence, and remove the least important tokens to preserve the predictive capability of the original network as much as possible. The experimental settings are in accordance with \cite{dynamic}. The pruning results are shown in Table \uppercase\expandafter{\romannumeral5}. For DeiT-S, we have achieved a 37.3\% reduction in FLOPs with only a 0.51\% loss in Top-1 accuracy and a 0.29\% loss in Top-5 accuracy. Similarly, for LVVIT-S, we have achieved a 42.1\% decrease in FLOPs, accompanied by a 1.36\% drop in Top-1 accuracy and a 0.56\% drop in Top-5 accuracy. This initial evidence strongly highlights the adaptability of our methodology with transformer-based models. Our future objective is to thoroughly explore advanced pruning techniques tailored to these models' unique characteristics.

\begin{table}[htbp]
\caption{Results of DeiT-S and LVVIT-S on ILSVRC-2012.\label{tab:table5}}
\centering
\setlength{\tabcolsep}{3mm}{
\begin{tabular}{cccc}
\toprule[1pt]
\multirow{1}{*}[-0.6ex]{Model} & \multirow{1}{*}[-0.6ex]{Top-1(\%)} & \multirow{1}{*}[-0.6ex]{GPU time(ms)} & \multirow{1}{*}[-0.6ex]{FLOPs[G] / PR[\%]} \\
\midrule 
DeiT-S \cite{deit} & 79.85 & 94.97 & 4.75 / 0.0 \\
Ours & 79.34 & 94.68 & 2.98 / 37.3 \\
\midrule 
LVVIT-S \cite{LVVIT} & 83.24 & 96.33 & 6.57 / 0.0 \\
Ours & 81.88 & 95.77 & 3.80 / 42.1 \\
\bottomrule[1pt] 
\end{tabular}}
\vspace{-0.5cm}
\end{table}

\vspace{-0.18cm}

\subsection{Results of Instance Segment on MS COCO}

Currently, the experiments we conducted fall under the category of image classification tasks. However, visual tasks also encompass objectives such as object detection, instance segmentation. To further validate the effectiveness of our method, we plan to apply pruning to YOLACT\cite{yolact}, a real-time instance segmentation model, on MS COCO\cite{mscoco}. For MS COCO, training is conducted on the train2017 dataset, and evaluation is performed on the val2017 and test-dev datasets. The backbone is set to ResNet-50-FPN. Fine-tuning has been conducted for 100 epochs, with a batch size of 16, while adhering to other settings as per YOLACT. The pruning results, detailed in Table \uppercase\expandafter{\romannumeral6}, reveal a notable reduction of 34.3\% in FLOPs, with only a marginal 0.97\% box mAP loss and 0.66\% mask mAP loss incurred. This provides strong evidence that our proposed method can achieve good results across a variety of visual tasks. In the future, we will explore the application of the proposed method in more vision tasks.

\begin{table}[htbp]
\caption{Results of YOLACT-ResNet50-FPN on MS COCO. The image size is 550×550. \label{tab:table6}}
\centering
\setlength{\tabcolsep}{2.1mm}{
\begin{tabular}{cccccc}
\toprule[1pt]
\multirow{1}{*}[-0.6ex]{Model} & \multirow{1}{*}[-0.6ex]{box mAP(\%)} & \multirow{1}{*}[-0.6ex]{mask mAP(\%)} & \multirow{1}{*}[-0.6ex]{FLOPs[G] / PR[\%]} \\
\midrule 
YOLACT\cite{yolact} & 28.40 & 26.75 & 59.29 / 0.0 \\
Ours &27.43  &26.09  & 38.97 / 34.3 \\
\bottomrule[1pt] 
\end{tabular}}
\vspace{-0.5cm}
\end{table}

\begin{figure}[htbp]
\centering
\includegraphics[width=.49\textwidth]{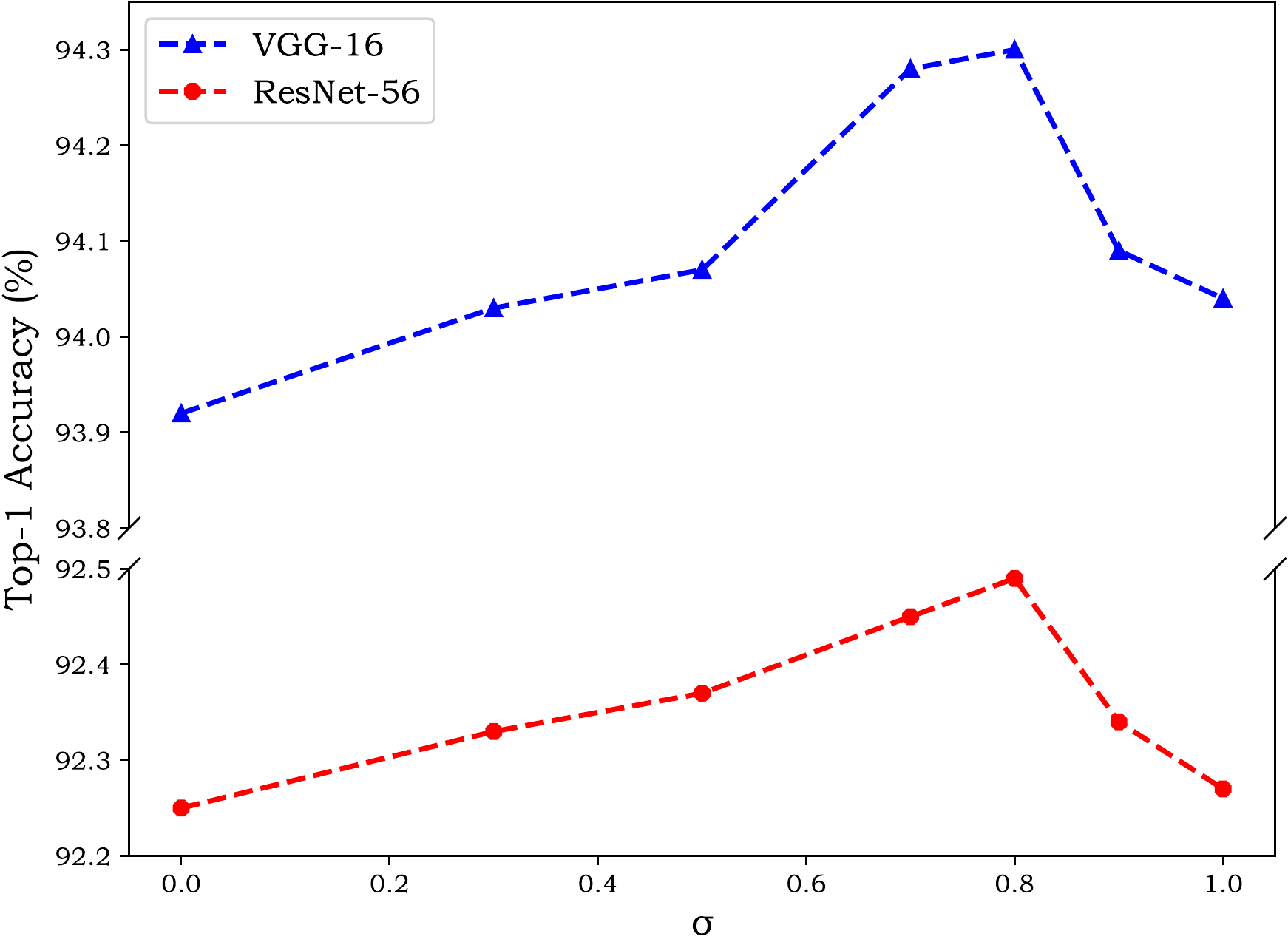}%
\captionsetup{justification=raggedright}
\caption{Top-1 Accuracy for variants of metric weight \( \sigma \). The upper figure is the accuracy of the pruned VGG-16, and the lower figure is about the pruned ResNet-56. The x-axis denotes different weight values, and the y-axis denotes the accuracy of the pruned model. Notably, the proposed method achieves optimal performance when the value of \( \sigma \) is within the range of 0.7 to 0.9.}
\label{fig_sim}
\vspace{-0.4cm}
\end{figure}

\begin{figure*}[!t]
\centering
\includegraphics[width=1.0\textwidth]{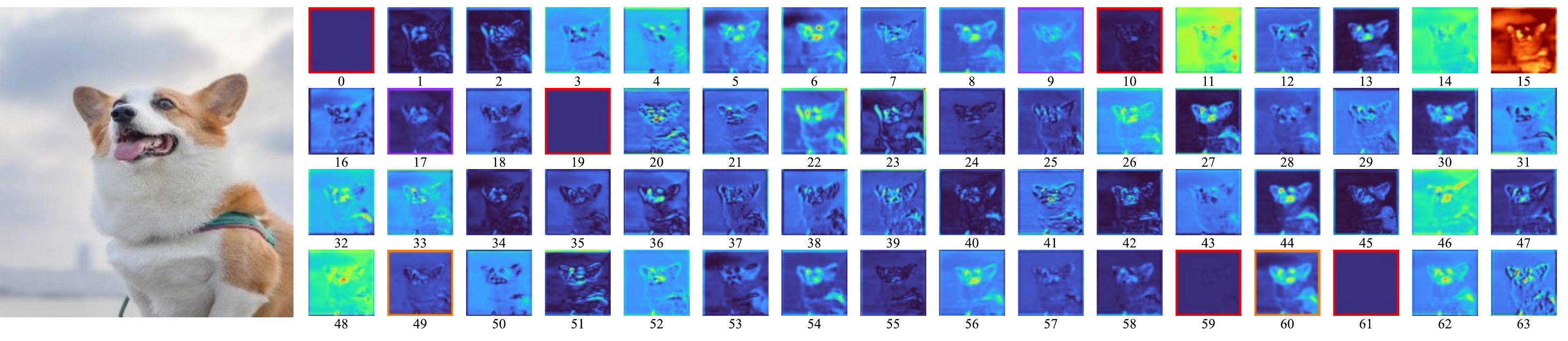}%
\captionsetup{justification=raggedright}
\caption{The graph depicts the feature maps of the intermediate layer output in the first block of ResNet-50. The input image is 224*224, the pruning rate is 10\% and the feature maps are numbered from 0 to 63, with filters corresponding one-to-one with the feature maps. We independently evaluate the importance of filters using information capacity and information independence. The purple box represents the least important filters exclusively selected by information capacity, corresponding to the indices (9, 17); the orange box represents the least important filters solely chosen by information independence, with indices (49, 60); and the red box represents filters simultaneously selected by both, with indices (0, 10, 19, 59, 61).}
\label{fig_sim}
\vspace{-0.55cm}
\end{figure*}

\vspace{-0.15cm} 

\subsection{Ablation Study}

\textit{1) Varying Metric Weight \( \sigma \):}
In this paper, we propose two metrics for evaluating the importance of filters: information capacity and information independence. Information capacity measures the information extraction capability of the filter, while information independence measures the correlations among filters. In other words, the former represents the local features inherent to the filters, and the latter represents the global features of the filters. We deem filters with high information capacity and independence to be important. To minimize the decline in predictive capability due to pruning, preserving the original network's information content is crucial. Thus, we argue that information capacity is a more critical metric than information independence. Using the weight \( \sigma \) to balance these metrics, we evaluate them on the VGG-16 and ResNet-56 architectures with the CIFAR-10 dataset for the purpose of compression. Fig. 4 displays the outcomes for various \( \sigma \) values. 
It is worth noting that the accuracy of both models initially increases as \( \sigma \) increases, but then begins to decline. The experimental results indicate that our method achieves optimal performance when the value of \( \sigma \) is within the range of 0.7 to 0.9. Consistently across all experiments detailed in this paper, we have set \( \sigma \) to 0.8. The experimental findings support our assertion that in a pruned network, retaining as much information as possible is more critical for a filter than maintaining information independence. Furthermore, it is evident that combining these two metrics outperforms the use of either metric in isolation (i.e., \( \sigma \) = 0.0 or 1.0). In other words, our proposed multi-perspective filter pruning approach offers distinct advantages in enhancing the performance of the pruned network over single-perspective methods.

\begin{figure}[!t]
\centering
\includegraphics[width=.48\textwidth]{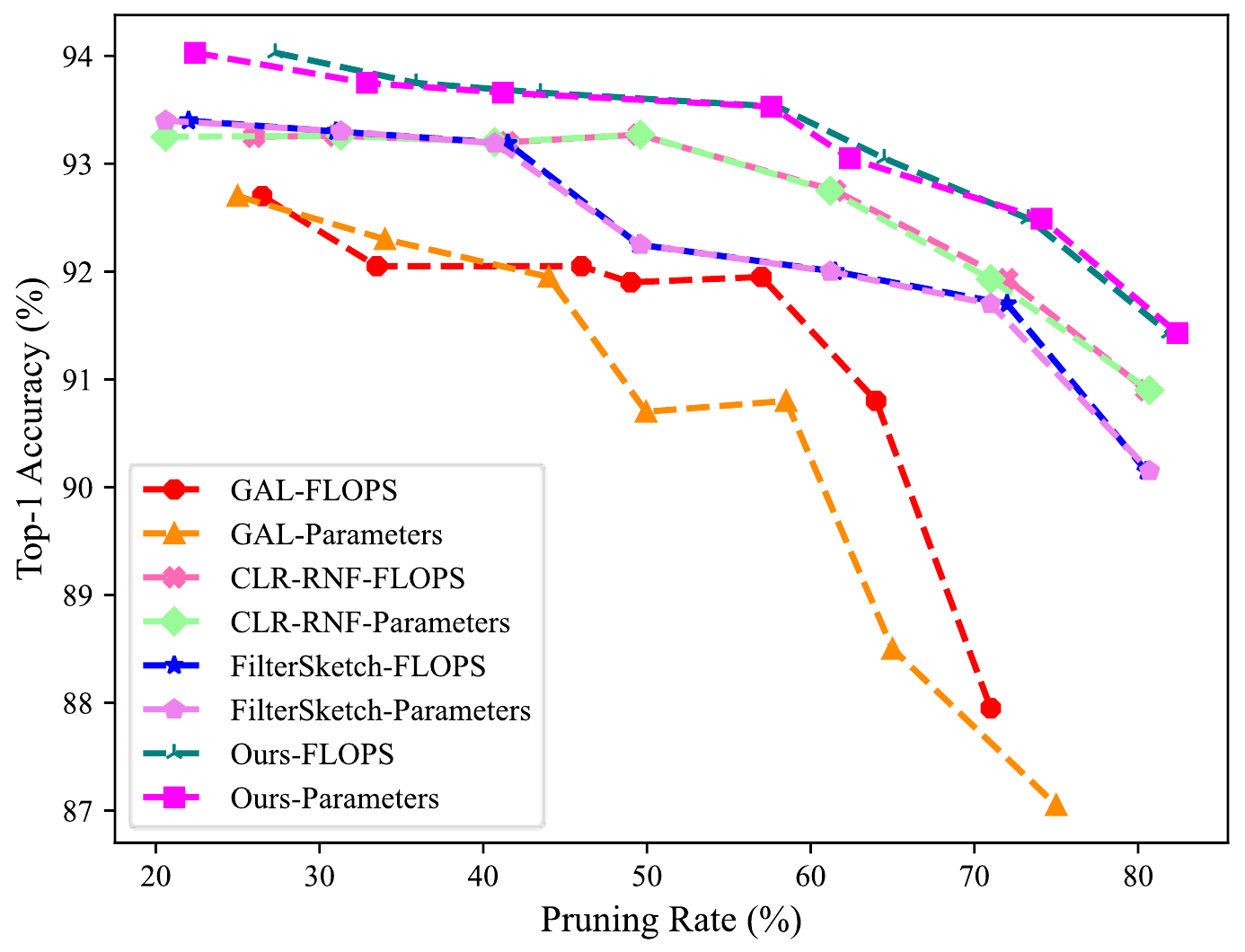}
\captionsetup{justification=raggedright}
\caption{Sensity analysis. Top-1 accuracy of ResNet-56 on CIFAR-10 with respect to diverse pruning rates. Different colored lines represent various pruning algorithms.}
\label{fig_1}
\vspace{-0.6cm}
\end{figure} 

\textit{2) Sensity Analysis:}
We evaluate the Top-1 accuracies and FLOPs reduction across various pruning rates for pruned networks on the ResNet-56 model, using the following methods: GAL \cite{lin2019towards}, CLR-RNF \cite{CLRRNF}, Filter-Sketch \cite{filtersketch}, and our approach. Notably, our approach exhibits significant improvements over the compared methods, as illustrated in Fig. 6. There is a significant drop in accuracy for GAL and Filter-Sketch as the pruning rate increases, in contrast, CLR-RNF and our approach maintain a relatively stable performance. Moreover, our approach consistently outperforms CLR-RNF at all measured points under similar compression ratios, further underscoring the effectiveness of our approach.

\textit{3) Varying Distance Types:}
To assess filter independence, we calculate the relative distance between one filter and the remaining filters using four distance metrics, including cosine distance and three kinds of minkowski distance. Experiments were conducted on ResNet-110, with a pruning rate of 71.9\% for FLOPs and 69.4\% for parameters. In Table \uppercase\expandafter{\romannumeral7} and \uppercase\expandafter{\romannumeral8}, ``Baseline'' and ``Pruned'' respectively represent the Top-1 accuracy for both the original and the pruned networks, and \( \Delta \) denotes the accuracy drop after pruning. As indicated in Table \uppercase\expandafter{\romannumeral7}, the highest accuracy of 93.78\% was achieved using the Euclidean distance metric. The accuracies based on Manhattan distance, Cosine distance and Chebyshev distance are 93.60\%, 93.40\% and 93.60\%, respectively. We argue that Euclidean distance is particularly effective for distinguishing the similarity among low-dimensional data points, such as 2D kernels, which are two-dimensional matrices. 

\begin{table}[htbp]
\caption{Results of ResNet-110 on CIFAR-10 with Different Distance Types. \label{tab:table7}}
\centering
\setlength{\tabcolsep}{3.7mm}{
\begin{tabular}{cccc}
\toprule[1pt]
\multirow{1}{*}[-0.6ex]{Distance Type} & \multirow{1}{*}[-0.6ex]{Baseline [\%]} & \multirow{1}{*}[-0.6ex]{Pruned [\%]} & \multirow{1}{*}[-0.6ex]{\( \Delta \) [\%]}\\
\midrule 
Euclidean Distance & 93.50 & 93.78 & 0.28  \\
Manhattan Distance & 93.50 & 93.60 & 0.10  \\
Chebyshev Distance & 93.50 & 93.40 & -0.02  \\
Cosine Distance & 93.50 & 93.60 & 0.10 \\
\bottomrule[1pt] 
\end{tabular}}
\vspace{-0.2cm}
\end{table}

\textit{4) Varying Pruning Strategies:}
To validate the efficacy of our proposed methodology comprehensively, we begin by computing the significance of each filter. Subsequently, we apply three distinct pruning strategies to ResNet-110, involving the removal of the least important, most important, and randomly selected filters. It is important to emphasize that, regardless of the specific pruning strategy employed, the pruning rates remain uniform across each layer. The results presented in Table \uppercase\expandafter{\romannumeral8} emphasize the superior performance achieved by our proposed importance metric. Notably, the highest Top-1 accuracy is consistently obtained when pruning the least important filters. This observation underscores the capability of our approach in discerning filters that exert minimal influence on the network, thus ensuring the preservation of maximum information and mitigating the associated accuracy decline resulting from the pruning process.
\begin{table}[htbp]
\caption{Results of ResNet-110 on CIFAR-10 with different pruning strategies, including pruning the least important, most important, and randomly selected filters.\label{tab:table8}}
\centering
\setlength{\tabcolsep}{4.2mm}{
\begin{tabular}{cccccc}
\toprule[1pt]
\multirow{1}{*}[-0.6ex]{Method} & \multirow{1}{*}[-0.6ex]{Baseline [\%]} & \multirow{1}{*}[-0.6ex]{Pruned [\%]} & \multirow{1}{*}[-0.6ex]{\( \Delta \) [\%]}\\
\midrule 
Least important & 93.50 & 93.78 & 0.28  \\
Most important & 93.50 & 93.38 & -0.12 \\
Random & 93.50 & 92.87 & -0.63 \\

\bottomrule[1pt] 
\end{tabular}}
\vspace{-0.3cm} 
\end{table}

\vspace{-0.2cm}

\subsection{Visualization of Feature Maps}
To gain a comprehensive understanding of the mechanism of the proposed method, we independently identify the least important filters using information capacity and information independence, as illustrated in Fig. 5. The strategy based on information capacity selects filters (9, 17), which correspond to feature maps with relatively vague features related to dogs and possess smaller information content. Hence, they are chosen by the information capacity-based strategy. On the other hand, the strategy based on information independence selects filters (49, 60), similar to filters (16, 18, 35, 22, 26, 62), and is thus chosen by the information independence-based strategy. It can be observed that this method effectively considers information at different levels of the network, maximizing the retention of information during the pruning process and consequently reducing the precision loss of the network. Furthermore, filters like (0, 10, 19, 59, 61), which have a relatively minor impact on the network, are also selected, indicating the effectiveness of this method in assessing filter importance.

\section{Conclusion}
This paper proposes a novel filter pruning approach using two metrics, namely information capacity and information independence, aiming to evaluate filters in an interpretable, multi-persepective, and lightweight manner. To this end, we leverage the interpretability and intuitiveness of information entropy and introduce it as a measure of the filter's information capacity. We empirically demonstrate that information entropy can capture relevant filter knowledge in a fine-grained manner. Our experimental results show that there is a clear correlation between feature map entropy and the corresponding filter, enabling us to compute the information capacity of the filter using only pre-trained weights. Additionally, we evaluate the impact of weight on model accuracy, in which this weight controls the balance between information capacity and information independence. The results reveal that our proposed multi-perspective filter pruning method outperforms single-perspective methods, including intra-channel and inter-channel methods, in improving the pruned network performance. Comprehensive experiments conducted on CIFAR-10/100 and ILSVRC-2012 substantiate the efficiency and effectiveness of our method in compressing and accelerating CNNs. Supplementary experiments on Transformer-like models and Instance Segmentation tasks provide additional validation of the efficacy of our approach. In our future research, we intend to investigate avenues for integrating our method with other model compression techniques, including but not limited to parameter quantization and knowledge distillation.

\normalem
\bibliographystyle{IEEEtran}
\bibliography{ref}

% Generated by IEEEtran.bst, version: 1.14 (2015/08/26)
\begin{thebibliography}{10}
\providecommand{\url}[1]{#1}
\csname url@samestyle\endcsname
\providecommand{\newblock}{\relax}
\providecommand{\bibinfo}[2]{#2}
\providecommand{\BIBentrySTDinterwordspacing}{\spaceskip=0pt\relax}
\providecommand{\BIBentryALTinterwordstretchfactor}{4}
\providecommand{\BIBentryALTinterwordspacing}{\spaceskip=\fontdimen2\font plus
\BIBentryALTinterwordstretchfactor\fontdimen3\font minus
  \fontdimen4\font\relax}
\providecommand{\BIBforeignlanguage}[2]{{%
\expandafter\ifx\csname l@#1\endcsname\relax
\typeout{** WARNING: IEEEtran.bst: No hyphenation pattern has been}%
\typeout{** loaded for the language `#1'. Using the pattern for}%
\typeout{** the default language instead.}%
\else
\language=\csname l@#1\endcsname
\fi
#2}}
\providecommand{\BIBdecl}{\relax}
\BIBdecl

\bibitem{krizhevsky2012imagenet}
A.~Krizhevsky, I.~Sutskever, and G.~E. Hinton, ``Imagenet classification with
  deep convolutional neural networks,'' in \emph{Proc. Adv. Neural Inf.
  Process. Syst. (NeurIPS)}, 2012.

\bibitem{he2016deep}
K.~He, X.~Zhang, S.~Ren, and J.~Sun, ``Deep residual learning for image
  recognition,'' in \emph{Proc. IEEE Conf. Comput. Vis. Pattern Recognit.
  (CVPR)}, 2016, pp. 770--778.

\bibitem{R2-Trans}
S.~Ye, S.~Yu, Y.~Wang, and X.~You, ``R2-trans: Fine-grained visual
  categorization with redundancy reduction,'' \emph{Image. Vis. Comput.}, vol.
  143, pp. 104\,923--104\,933, 2024.

\bibitem{ye2022discriminative}
S.~Ye, Q.~Peng, W.~Sun, J.~Xu, Y.~Wang, X.~You, and Y.-M. Cheung,
  ``Discriminative suprasphere embedding for fine-grained visual
  categorization,'' \emph{IEEE Trans. Neural Netw. Learn. Syst.}, 2022.

\bibitem{simonyan2014very}
K.~Simonyan and A.~Zisserman, ``Very deep convolutional networks for
  large-scale image recognition,'' \emph{Proc. Int. Conf. Learn. Rep. (ICLR)},
  2015.

\bibitem{russakovsky2015imagenet}
O.~Russakovsky, J.~Deng, H.~Su, J.~Krause, S.~Satheesh, S.~Ma, Z.~Huang,
  A.~Karpathy, A.~Khosla, M.~Bernstein \emph{et~al.}, ``Imagenet large scale
  visual recognition challenge,'' \emph{Int. J. Comput. Vis.}, vol. 115, no.~3,
  pp. 211--252, 2015.

\bibitem{zhang2018shufflenet}
X.~Zhang, X.~Zhou, M.~Lin, and J.~Sun, ``Shufflenet: An extremely efficient
  convolutional neural network for mobile devices,'' in \emph{Proc. IEEE Conf.
  Comput. Vis. Pattern Recognit. (CVPR)}, 2018, pp. 6848--6856.

\bibitem{girshick2015fast}
R.~Girshick, ``Fast r-cnn,'' in \emph{Proc. IEEE Int. Conf. Comput. Vis.
  (ICCV)}, 2015, pp. 1440--1448.

\bibitem{redmon2016you}
J.~Redmon, S.~Divvala, R.~Girshick, and A.~Farhadi, ``You only look once:
  Unified, real-time object detection,'' in \emph{Proc. IEEE Conf. Comput. Vis.
  Pattern Recognit. (CVPR)}, 2016, pp. 779--788.

\bibitem{ren2015faster}
S.~Ren, K.~He, R.~Girshick, and J.~Sun, ``Faster r-cnn: Towards real-time
  object detection with region proposal networks,'' in \emph{Proc. Adv. Neural
  Inf. Process. Syst. (NeurIPS)}, 2015.

\bibitem{liu2016ssd}
W.~Liu, D.~Anguelov, D.~Erhan, C.~Szegedy, S.~Reed, C.-Y. Fu, and A.~C. Berg,
  ``Ssd: Single shot multibox detector,'' in \emph{Proc. Eur. Conf. Comput.
  Vis. (ECCV)}, 2016, pp. 21--37.

\bibitem{zhu2021semantic}
C.~Zhu, F.~Chen, U.~Ahmed, Z.~Shen, and M.~Savvides, ``Semantic relation
  reasoning for shot-stable few-shot object detection,'' in \emph{Proc.
  IEEE/CVF Conf. Comput. Vis. Pattern Recognit. (CVPR)}, 2021, pp. 8782--8791.

\bibitem{yan2018spatial}
S.~Yan, Y.~Xiong, and D.~Lin, ``Spatial temporal graph convolutional networks
  for skeleton-based action recognition,'' in \emph{Proc. AAAI Conf. Artif.
  Intelli. (AAAI)}, 2018.

\bibitem{li2019actional}
M.~Li, S.~Chen, X.~Chen, Y.~Zhang, Y.~Wang, and Q.~Tian, ``Actional-structural
  graph convolutional networks for skeleton-based action recognition,'' in
  \emph{Proc. IEEE/CVF Conf. Comput. Vis. Pattern Recognit. (CVPR)}, 2019, pp.
  3595--3603.

\bibitem{tran2015learning}
D.~Tran, L.~Bourdev, R.~Fergus, L.~Torresani, and M.~Paluri, ``Learning
  spatiotemporal features with 3d convolutional networks,'' in \emph{Proc. IEEE
  Int. Conf. Comput. Vis. (ICCV)}, 2015, pp. 4489--4497.

\bibitem{qiu2017learning}
Z.~Qiu, T.~Yao, and T.~Mei, ``Learning spatio-temporal representation with
  pseudo-3d residual networks,'' in \emph{Proc. IEEE Int. Conf. Comput. Vis.
  (ICCV)}, 2017, pp. 5533--5541.

\bibitem{feichtenhofer2016convolutional}
C.~Feichtenhofer, A.~Pinz, and A.~Zisserman, ``Convolutional two-stream network
  fusion for video action recognition,'' in \emph{Proc. IEEE Conf. Comput. Vis.
  Pattern Recognit. (CVPR)}, 2016, pp. 1933--1941.

\bibitem{chen2013cooperative}
G.~Chen and Y.-D. Song, ``Cooperative tracking control of nonlinear multiagent
  systems using self-structuring neural networks,'' \emph{IEEE Trans. Neural
  Netw. Learn. Syst.}, vol.~25, no.~8, pp. 1496--1507, 2013.

\bibitem{chen2016terminal}
G.~Chen, Y.~Song, and Y.~Guan, ``Terminal sliding mode-based consensus tracking
  control for networked uncertain mechanical systems on digraphs,'' \emph{IEEE
  Trans. Neural Netw. Learn. Syst.}, vol.~29, no.~3, pp. 749--756, 2016.

\bibitem{han2015learning}
S.~Han, J.~Pool, J.~Tran, and W.~Dally, ``Learning both weights and connections
  for efficient neural network,'' in \emph{Proc. Adv. Neural Inf. Process.
  Syst. (NeurIPS)}, 2015.

\bibitem{luo2017thinet}
J.-H. Luo, J.~Wu, and W.~Lin, ``Thinet: A filter level pruning method for deep
  neural network compression,'' in \emph{Proc. IEEE Int. Conf. Comput. Vis.
  (ICCV)}, 2017, pp. 5058--5066.

\bibitem{taylor}
P.~Molchanov, A.~Mallya, S.~Tyree, I.~Frosio, and J.~Kautz, ``Importance
  estimation for neural network pruning,'' in \emph{Proc. IEEE/CVF Conf.
  Comput. Vis. Pattern Recognit. (CVPR)}, 2019, pp. 11\,264--11\,272.

\bibitem{li2020group}
Y.~Li, S.~Gu, C.~Mayer, L.~V. Gool, and R.~Timofte, ``Group sparsity: The hinge
  between filter pruning and decomposition for network compression,'' in
  \emph{Proc. IEEE/CVF Conf. Comput. Vis. Pattern Recognit. (CVPR)}, 2020, pp.
  8018--8027.

\bibitem{filtersketch}
M.~Lin, L.~Cao, S.~Li, Q.~Ye, Y.~Tian, J.~Liu, Q.~Tian, and R.~Ji, ``Filter
  sketch for network pruning,'' \emph{IEEE Trans. Neural Netw. Learn. Syst.},
  2021.

\bibitem{zhang2015efficient}
X.~Zhang, J.~Zou, X.~Ming, K.~He, and J.~Sun, ``Efficient and accurate
  approximations of nonlinear convolutional networks,'' in \emph{Proc. IEEE
  Conf. Comput. Vis. Pattern Recognit. (CVPR)}, 2015, pp. 1984--1992.

\bibitem{yin2021towards}
M.~Yin, Y.~Sui, S.~Liao, and B.~Yuan, ``Towards efficient tensor
  decomposition-based dnn model compression with optimization framework,'' in
  \emph{Proc. IEEE/CVF Conf. Comput. Vis. Pattern Recognit. (CVPR)}, 2021, pp.
  10\,674--10\,683.

\bibitem{lin2018holistic}
S.~Lin, R.~Ji, C.~Chen, D.~Tao, and J.~Luo, ``Holistic cnn compression via
  low-rank decomposition with knowledge transfer,'' \emph{IEEE Trans. Pattern
  Anal. Mach. Intell.}, vol.~41, no.~12, pp. 2889--2905, 2018.

\bibitem{hayashi2019exploring}
K.~Hayashi, T.~Yamaguchi, Y.~Sugawara, and S.-i. Maeda, ``Exploring unexplored
  tensor network decompositions for convolutional neural networks,'' in
  \emph{Proc. Adv. Neural Inf. Process. Syst. (NeurIPS)}, 2019.

\bibitem{lin2020rotated}
M.~Lin, R.~Ji, Z.~Xu, B.~Zhang, Y.~Wang, Y.~Wu, F.~Huang, and C.-W. Lin,
  ``Rotated binary neural network,'' in \emph{Proc. Adv. Neural Inf. Process.
  Syst. (NeurIPS)}, 2020, pp. 7474--7485.

\bibitem{liu2020bi}
Z.~Liu, W.~Luo, B.~Wu, X.~Yang, W.~Liu, and K.-T. Cheng, ``Bi-real net:
  Binarizing deep network towards real-network performance,'' \emph{Int. J.
  Comput. Vis.}, vol. 128, pp. 202--219, 2020.

\bibitem{tung2019similarity}
F.~Tung and G.~Mori, ``Similarity-preserving knowledge distillation,'' in
  \emph{Proc. IEEE/CVF Int. Conf. Comput. Vis.}, 2019, pp. 1365--1374.

\bibitem{park2019relational}
W.~Park, D.~Kim, Y.~Lu, and M.~Cho, ``Relational knowledge distillation,'' in
  \emph{Proc. IEEE/CVF Conf. Comput. Vis. Pattern Recognit. (CVPR)}, 2019, pp.
  3967--3976.

\bibitem{liu2018frequency}
Z.~Liu, J.~Xu, X.~Peng, and R.~Xiong, ``Frequency-domain dynamic pruning for
  convolutional neural networks,'' in \emph{Proc. Adv. Neural Inf. Process.
  Syst. (NeurIPS)}, 2018.

\bibitem{chen2018constraint}
C.~Chen, F.~Tung, N.~Vedula, and G.~Mori, ``Constraint-aware deep neural
  network compression,'' in \emph{Proc. Eur. Conf. Comput. Vis. (ECCV)}, 2018,
  pp. 400--415.

\bibitem{BLIS1}
\BIBentryALTinterwordspacing
F.~G. {V}an {Z}ee and R.~A. {v}an~{d}e {G}eijn, ``{BLIS}: A framework for
  rapidly instantiating {BLAS} functionality,'' \emph{ACM Trans. Math. Softw.},
  vol.~41, no.~3, pp. 14:1--14:33, 2015. [Online]. Available:
  \url{https://doi.acm.org/10.1145/2764454}
\BIBentrySTDinterwordspacing

\bibitem{he2017channel}
Y.~He, X.~Zhang, and J.~Sun, ``Channel pruning for accelerating very deep
  neural networks,'' in \emph{Proc. IEEE Int. Conf. Comput. Vis. (ICCV)}, 2017,
  pp. 1389--1397.

\bibitem{yu2018nisp}
R.~Yu, A.~Li, C.-F. Chen, J.-H. Lai, V.~I. Morariu, X.~Han, M.~Gao, C.-Y. Lin,
  and L.~S. Davis, ``Nisp: Pruning networks using neuron importance score
  propagation,'' in \emph{Proc. IEEE Conf. Comput. Vis. Pattern Recognit.
  (CVPR)}, 2018, pp. 9194--9203.

\bibitem{sfp}
Y.~He, G.~Kang, X.~Dong, Y.~Fu, and Y.~Yang, ``Soft filter pruning for
  accelerating deep convolutional neural networks,'' in \emph{Int. Joint Conf.
  Artif, Intelli. (IJCAI)}, 2018, pp. 2234--2240.

\bibitem{guo2021gdp}
Y.~Guo, H.~Yuan, J.~Tan, Z.~Wang, S.~Yang, and J.~Liu, ``Gdp: Stabilized neural
  network pruning via gates with differentiable polarization,'' in \emph{Proc.
  IEEE/CVF Int. Conf. Comput. Vis. (ICCV)}, 2021, pp. 5239--5250.

\bibitem{liu2019metapruning}
Z.~Liu, H.~Mu, X.~Zhang, Z.~Guo, X.~Yang, K.-T. Cheng, and J.~Sun,
  ``Metapruning: Meta learning for automatic neural network channel pruning,''
  in \emph{Proc. IEEE/CVF Int. Conf. Comput. Vis. (ICCV)}, 2019, pp.
  3296--3305.

\bibitem{li2016pruning}
H.~Li, A.~Kadav, I.~Durdanovic, H.~Samet, and H.~P. Graf, ``Pruning filters for
  efficient convnets,'' in \emph{Proc. Int. Conf. Learn. Rep. (ICLR)}, 2017,
  pp. 1--9.

\bibitem{lin2020hrank}
M.~Lin, R.~Ji, Y.~Wang, Y.~Zhang, B.~Zhang, Y.~Tian, and L.~Shao, ``Hrank:
  Filter pruning using high-rank feature map,'' in \emph{Proc. IEEE/CVF Conf.
  Comput. Vis. Pattern Recognit. (CVPR)}, 2020, pp. 1529--1538.

\bibitem{he2019filter}
Y.~He, P.~Liu, Z.~Wang, Z.~Hu, and Y.~Yang, ``Filter pruning via geometric
  median for deep convolutional neural networks acceleration,'' in \emph{Proc.
  IEEE/CVF Conf. Comput. Vis. Pattern Recognit. (CVPR)}, 2019, pp. 4340--4349.

\bibitem{sui2021chip}
Y.~Sui, M.~Yin, Y.~Xie, H.~Phan, S.~Aliari~Zonouz, and B.~Yuan, ``Chip: Channel
  independence-based pruning for compact neural networks,'' in \emph{Proc. Adv.
  Neural Inf. Process. Syst. (NeurIPS)}, 2021, pp. 24\,604--24\,616.

\bibitem{li2019exploiting}
Y.~Li, S.~Lin, B.~Zhang, J.~Liu, D.~Doermann, Y.~Wu, F.~Huang, and R.~Ji,
  ``Exploiting kernel sparsity and entropy for interpretable cnn compression,''
  in \emph{Proc. IEEE/CVF Conf. Comput. Vis. Pattern Recognit. (CVPR)}, 2019,
  pp. 2800--2809.

\bibitem{hu2016network}
H.~Hu, R.~Peng, Y.-W. Tai, and C.-K. Tang, ``Network trimming: A data-driven
  neuron pruning approach towards efficient deep architectures,'' \emph{arXiv
  preprint arXiv:1607.03250}, 2016.

\bibitem{wang2018exploring}
D.~Wang, L.~Zhou, X.~Zhang, X.~Bai, and J.~Zhou, ``Exploring linear
  relationship in feature map subspace for convnets compression,'' \emph{arXiv
  preprint arXiv:1803.05729}, 2018.

\bibitem{dubey2018coreset}
A.~Dubey, M.~Chatterjee, and N.~Ahuja, ``Coreset-based neural network
  compression,'' in \emph{Proc. Eur. Conf. Comput. Vis. (ECCV)}, 2018, pp.
  454--470.

\bibitem{tang2020scop}
Y.~Tang, Y.~Wang, Y.~Xu, D.~Tao, C.~Xu, C.~Xu, and C.~Xu, ``Scop: Scientific
  control for reliable neural network pruning,'' in \emph{Proc. Adv. Neural
  Inf. Process. Syst. (NeurIPS)}, 2020, pp. 10\,936--10\,947.

\bibitem{peng2019collaborative}
H.~Peng, J.~Wu, S.~Chen, and J.~Huang, ``Collaborative channel pruning for deep
  networks,'' in \emph{Int. Conf. Mach. Learn. (ICML)}, 2019, pp. 5113--5122.

\bibitem{tiwari2021chipnet}
R.~Tiwari, U.~Bamba, A.~Chavan, and D.~K. Gupta, ``Chipnet: Budget-aware
  pruning with heaviside continuous approximations,'' \emph{arXiv preprint
  arXiv:2102.07156}, 2021.

\bibitem{lemaire2019structured}
C.~Lemaire, A.~Achkar, and P.-M. Jodoin, ``Structured pruning of neural
  networks with budget-aware regularization,'' in \emph{Proc. IEEE/CVF Conf.
  Comput. Vis. Pattern Recognit. (CVPR)}, 2019, pp. 9108--9116.

\bibitem{zhuo2018scsp}
H.~Zhuo, X.~Qian, Y.~Fu, H.~Yang, and X.~Xue, ``Scsp: Spectral clustering
  filter pruning with soft self-adaption manners,'' \emph{arXiv preprint
  arXiv:1806.05320}, 2018.

\bibitem{ye2018rethinking}
J.~Ye, X.~Lu, Z.~Lin, and J.~Z. Wang, ``Rethinking the
  smaller-norm-less-informative assumption in channel pruning of convolution
  layers,'' in \emph{Proc. Int. Conf. Learn. Rep. (ICLR)}, 2018.

\bibitem{suau2018principal}
X.~Suau, L.~Zappella, V.~Palakkode, and N.~Apostoloff, ``Principal filter
  analysis for guided network compression,'' \emph{arXiv preprint
  arXiv:1807.10585}, vol.~2, 2018.

\bibitem{yu2020understanding}
S.~Yu, K.~Wickstr{\o}m, R.~Jenssen, and J.~C. Principe, ``Understanding
  convolutional neural networks with information theory: An initial
  exploration,'' \emph{IEEE Trans. Neural Netw. Learn. Syst.}, vol.~32, no.~1,
  pp. 435--442, 2020.

\bibitem{Yu2020Multivariate}
S.~Yu, L.~G.~S. Giraldo, R.~Jenssen, and J.~C. Príncipe, ``Multivariate
  extension of matrix-based rényi's $\alpha$-order entropy functional,''
  \emph{IEEE Trans. Pattern Anal. Mach. Intell.}, vol.~42, no.~11, pp.
  2960--2966, 2020.

\bibitem{shi2022deep}
Y.~Shi, S.~Chen, X.~You, Q.~Peng, W.~Ou, and Y.~Zhao, ``Deep supervised
  information bottleneck hashing for cross-modal retrieval based computer-aided
  diagnosis,'' in \emph{Proc. AAAI Workshop Inf. Theory Deep Learn. (IT4DL)},
  2022.

\bibitem{ye2023coping}
S.~Ye, S.~Yu, W.~Hou, Y.~Wang, and X.~You, ``Coping with change: Learning
  invariant and minimum sufficient representations for fine-grained visual
  categorization,'' \emph{Comput. Vis. Image Underst.}, vol. 237, pp.
  103\,837--103\,870, 2023.

\bibitem{min20182pfpce}
C.~Min, A.~Wang, Y.~Chen, W.~Xu, and X.~Chen, ``2pfpce: Two-phase filter
  pruning based on conditional entropy,'' in \emph{Proc. AAAI Conf. Artif.
  Intelli. (AAAI)}, 2018.

\bibitem{ganesh2021mint}
M.~R. Ganesh, J.~J. Corso, and S.~Y. Sekeh, ``Mint: Deep network compression
  via mutual information-based neuron trimming,'' in \emph{Int. Conf. Pattern
  Recognit. (ICPR)}, 2021, pp. 8251--8258.

\bibitem{hrel}
C.~Sarvani, M.~Ghorai, S.~R. Dubey, and S.~S. Basha, ``Hrel: Filter pruning
  based on high relevance between activation maps and class labels,''
  \emph{Neural Netw.}, vol. 147, pp. 186--197, 2022.

\bibitem{principe2010information}
J.~C. Principe, \emph{Information theoretic learning: Rényi's entropy and
  kernel perspectives}.\hskip 1em plus 0.5em minus 0.4em\relax Springer Science
  \& Business Media, 2010.

\bibitem{krizhevsky2009learning}
A.~Krizhevsky and G.~Hinton, ``Learning multiple layers of features from tiny
  images,'' \emph{Handbook of Systemic Autoimmune Diseases}, vol.~1, no.~4,
  2009.

\bibitem{mscoco}
T.-Y. Lin, M.~Maire, S.~Belongie, J.~Hays, P.~Perona, D.~Ramanan,
  P.~Doll{\'a}r, and C.~L. Zitnick, ``Microsoft coco: Common objects in
  context,'' in \emph{Proc. Eur. Conf. Comput. Vis. (ECCV)}, 2014, pp.
  740--755.

\bibitem{huang2018data}
Z.~Huang and N.~Wang, ``Data-driven sparse structure selection for deep neural
  networks,'' in \emph{Proc. Eur. Conf. Comput. Vis. (ECCV)}, 2018, pp.
  304--320.

\bibitem{lin2019towards}
S.~Lin, R.~Ji, C.~Yan, B.~Zhang, L.~Cao, Q.~Ye, F.~Huang, and D.~Doermann,
  ``Towards optimal structured cnn pruning via generative adversarial
  learning,'' in \emph{Proc. IEEE/CVF Conf. Comput. Vis. Pattern Recognit.
  (CVPR)}, 2019, pp. 2790--2799.

\bibitem{switch}
Y.~He, P.~Liu, L.~Zhu, and Y.~Yang, ``Filter pruning by switching to
  neighboring cnns with good attributes,'' \emph{IEEE Trans. Neural Netw.
  Learn. Syst.}, vol.~34, no.~10, pp. 8044--8056, 2022.

\bibitem{Coarse-to-Fine}
S.~Lee and B.~C. Song, ``Fast filter pruning via coarse-to-fine neural
  architecture search and contrastive knowledge transfer,'' \emph{IEEE Trans.
  Neural Netw. Learn. Syst.}, 2023.

\bibitem{FSIM-E}
Y.~Liu, K.~Fan, D.~Wu, and W.~Zhou, ``Filter pruning by quantifying feature
  similarity and entropy of feature maps,'' \emph{Neurocomputing}, vol. 544,
  pp. 126\,297--126\,308, 2023.

\bibitem{rgp}
Z.~Chen, J.~Xiang, Y.~Lu, Q.~Xuan, Z.~Wang, G.~Chen, and X.~Yang, ``Rgp: Neural
  network pruning through regular graph with edges swapping,'' \emph{IEEE
  Trans. Neural Netw. Learn. Syst.}, 2023.

\bibitem{CLRRNF}
M.~Lin, L.~Cao, Y.~Zhang, L.~Shao, C.-W. Lin, and R.~Ji, ``Pruning networks
  with cross-layer ranking \& k-reciprocal nearest filters,'' \emph{IEEE Trans.
  Neural Netw. Learn. Syst.}, vol.~34, no.~11, pp. 9139--9148, 2022.

\bibitem{white-box}
Y.~Zhang, M.~Lin, C.-W. Lin, J.~Chen, Y.~Wu, Y.~Tian, and R.~Ji, ``Carrying out
  cnn channel pruning in a white box,'' \emph{IEEE Trans. Neural Netw. Learn.
  Syst.}, vol.~34, no.~10, pp. 7946--7955, 2022.

\bibitem{paszke2017automatic}
A.~Paszke, S.~Gross, S.~Chintala, G.~Chanan, E.~Yang, Z.~DeVito, Z.~Lin,
  A.~Desmaison, L.~Antiga, and A.~Lerer, ``Automatic differentiation in
  pytorch,'' 2017.

\bibitem{vit}
A.~Dosovitskiy, L.~Beyer, A.~Kolesnikov, D.~Weissenborn, X.~Zhai,
  T.~Unterthiner, M.~Dehghani, M.~Minderer, G.~Heigold, S.~Gelly \emph{et~al.},
  ``An image is worth 16x16 words: Transformers for image recognition at
  scale,'' in \emph{Proc. Int. Conf. Learn. Rep. (ICLR)}, 2021.

\bibitem{swin}
Z.~Liu, Y.~Lin, Y.~Cao, H.~Hu, Y.~Wei, Z.~Zhang, S.~Lin, and B.~Guo, ``Swin
  transformer: Hierarchical vision transformer using shifted windows,'' in
  \emph{Proc. IEEE/CVF Int. Conf. Comput. Vis. (ICCV)}, 2021, pp.
  10\,012--10\,022.

\bibitem{deit}
H.~Touvron, M.~Cord, M.~Douze, F.~Massa, A.~Sablayrolles, and H.~J{\'e}gou,
  ``Training data-efficient image transformers \& distillation through
  attention,'' in \emph{Int. Conf. Mach. Learn. (ICML)}, 2021, pp.
  10\,347--10\,357.

\bibitem{LVVIT}
Z.~Jiang, Q.~Hou, L.~Yuan, D.~Zhou, X.~Jin, A.~Wang, and J.~Feng, ``Token
  labeling: Training a 85.5\% top-1 accuracy vision transformer with 56m
  parameters on imagenet,'' \emph{arXiv preprint arXiv:2104.10858}, 2021.

\bibitem{dynamic}
Y.~Rao, W.~Zhao, B.~Liu, J.~Lu, J.~Zhou, and C.-J. Hsieh, ``Dynamicvit:
  Efficient vision transformers with dynamic token sparsification,'' in
  \emph{Proc. Adv. Neural Inf. Process. Syst. (NeurIPS)}, 2021, pp.
  13\,937--13\,949.

\bibitem{yolact}
D.~Bolya, C.~Zhou, F.~Xiao, and Y.~J. Lee, ``Yolact: Real-time instance
  segmentation,'' in \emph{Proc. IEEE/CVF Int. Conf. Comput. Vis. (ICCV)},
  2019, pp. 9157--9166.

\end{thebibliography}

\vfill

\end{document}